\documentclass[]{fairmeta}
\usepackage[utf8]{inputenc}
\usepackage[T1]{fontenc} 
\usepackage{hyperref}   
\usepackage{url}        
\usepackage{booktabs}       
\usepackage{amsfonts}       
\usepackage{nicefrac}      
\usepackage{microtype}     
\usepackage{xcolor}  
\definecolor{md-bg}{HTML}{F5F5F5} % light gray background
\usepackage{tikz}
\usepackage{algorithm}
\usepackage[noend]{algpseudocode}
\usepackage{listings}
\usepackage{caption}
\usepackage{amsthm}
\usepackage{amsmath}
\usepackage[inline]{enumitem}
\usepackage{xspace} 
\usepackage{cleveref}
\usepackage{subscript}
\usepackage{fvextra}
\usepackage{framed}
\usepackage{color}
\usepackage{graphicx} 
\usepackage{wrapfig} 
\usepackage{subcaption}
\usepackage{mathtools}
\usepackage{adjustbox}
\usepackage{siunitx}
\usepackage{twemojis}
\usepackage{tabularx}
\usepackage{fontawesome5}
\usepackage{amssymb}

\usepackage[labelsep=colon]{caption}
\usepackage[normalem]{ulem}  
\definecolor{shadecolor}{rgb}{0.95, 0.95, 0.95}

\algrenewcommand\algorithmicrequire{\textbf{Require:}}
\algrenewcommand\algorithmicensure{\textbf{Ensure:}}

\definecolor{lightgray}{gray}{0.95}
\definecolor{darkblue}{rgb}{0.1,0.2,0.6}
\definecolor{codegray}{rgb}{0.4,0.4,0.4}

\usetikzlibrary{shapes.geometric, arrows.meta, positioning, fit, calc, backgrounds, shadows}

\definecolor{primaryBlue}{RGB}{0, 105, 148}
\definecolor{workerGreen}{RGB}{46, 139, 87}
\definecolor{evalOrange}{RGB}{230, 126, 34}
\definecolor{outputGold}{RGB}{241, 196, 15}
\definecolor{darkGray}{RGB}{50, 50, 50}

\newcommand{\draftnote}[3]{\textcolor{#1}{[#2:~#3]}}

\newcommand{\ag}[1]{\draftnote{magenta}{AlexG}{#1}} % example: Alex Goldie
\usepackage{natbib}

\usepackage{xcolor}
\usepackage{tikz}
\usetikzlibrary{positioning, arrows.meta, calc}

\definecolor{lvlone}{RGB}{31, 119, 180}
\definecolor{lvltwo}{RGB}{214, 97, 43}
\definecolor{lvlthree}{RGB}{117, 82, 160}



\lstdefinestyle{pddlstyle}{
  numbers=left,
  numbersep=-6pt,
  backgroundcolor=\color{lightgray},
  basicstyle=\ttfamily\scriptsize,
  keywordstyle=\color{darkblue}\bfseries,
  commentstyle=\color{codegray}\itshape,
  showstringspaces=false,
  frame=single,
  breaklines=true,
  language=PDDL
}

\lstdefinestyle{nl}{
  numbers=left,
  numbersep=-6pt,
  backgroundcolor=\color{lightgray},
  basicstyle=\ttfamily\scriptsize,
  keywordstyle=\color{darkblue}\bfseries,
  commentstyle=\color{codegray}\itshape,
  showstringspaces=false,
  frame=single,
  breaklines=true,
  breakindent=0pt, 
  mathescape=true,
  lineskip=-1pt, 
}

\lstdefinestyle{pythonstyle}{
  numbers=none,
  numbersep=-6pt,
  backgroundcolor=\color{lightgray},
  basicstyle=\ttfamily\scriptsize,
  keywordstyle=\color{darkblue}\bfseries,
  commentstyle=\color{codegray}\itshape,
  stringstyle=\color{codegray},
  showstringspaces=false,
  frame=single,
  breaklines=true,
  breakindent=0pt,
  language=Python,
}

\usetikzlibrary{shapes.misc}

\crefname{figure}{Fig.}{Figures}

\newcommand{\airadojo}{{AIRA-dojo}\xspace}

\newcommand{\improve}{{\textsc{Improve}}\xspace}
\newcommand{\draft}{{\textsc{Draft}}\xspace}
\newcommand{\debug}{{\textsc{Debug}}\xspace}

\newcommand{\atlas}{{\textsc{AIRA\textsubscript{2}}}\xspace}

\newcommand{\airsbench}{\textsc{AIRS-Bench}}

\newcommand{\mlgym}{\textsc{MLGym}}

\title{\parbox{\linewidth}{\centering AI Research Preference Models}\vspace{.5em}}

\author[1,2,\dagger]{Thomas Simon Foster}
\author[1, \dagger]{Bassel Al Omari}
\author[1,2,\dagger]{Tingchen Fu}
\author[1,*]{Thomas Mann}
\author[1,*]{Carl Domond}
\author[1,]{\\Lucia Cipolina-Kun}
\author[1,]{Bhavul Gauri}
\author[1,]{Muna Aghamelu}
\author[1, 2,]{Alexander D. Goldie}
\author[1,]{Eryk Helenowski}
\author[1]{Jean-Christophe Gagnon-Audet}
\author[1]{Alberto Pepe}
\author[1]{Saba Nazir}
\author[1]{Daniel Izcovich}
\author[1]{Noam Levi}
\author[1]{Rishi Hazra}
\author[1, 3]{Karen Hambardzumyan}
\author[1]{Nicolas Baldwin}
\author[1]{Xian Li}
\author[1]{Martin Josifoski}
\author[1]{Paris Giampouras}
\author[1]{Masoud Jalili Sabet}
\author[1]{Anya Sims}
\author[1]{Hela Momand}
\author[1]{Tatiana Shavrina}
\author[1]{Despoina Magka}

\author[1]{Jason Weston}
\author[2]{Yulin Wang}
\author[1]{Anirudh Goyal}
\author[2]{João Henriques}
\author[1]{Yoram Bachrach}
\author[1, \ddagger]{Emily McMilin}
\author[1, 2, \ddagger]{Jakob Nicolaus Foerster}

\affiliation[1]{FAIR at Meta}
\affiliation[2]{University of Oxford}
\affiliation[3]{University College London}

\contribution[\dagger]{Lead Authors}
\contribution[*]{Core Contribution}
\contribution[\ddagger]{Equal Supervision}

\abstract{AI research agents (AIRA) can now carry machine learning experiments from proposal through implementation and evaluation. Yet progress on frontier tasks is throttled by the cost of evaluations that can consume days of GPU time. When an agent can propose far more candidates than it can afford to run, progress depends on its \textit{research preference}: how it allocates a fixed execution budget across many candidates. We introduce AI Research Preference Models (RPMs) that predict which candidate solution is most promising, without paying the cost of running them all. We build RPMs from frozen pretrained language models in two variants: an inference-only model that reasons over candidate plans, code, and previously executed solutions, and an agentic model that additionally runs small-scale pilot experiments. Integrated into the \airadojo{} research agent and evaluated on the machine learning research benchmark \airsbench{}, the two variants increase the average normalized score from $0.684$ to $0.711$ and $0.729$, respectively. Both reach the unguided agent's 24-hour performance in roughly 15 hours, using less than two-thirds of its execution budget, and together yield new state-of-the-art results on two \airsbench{} tasks.}
\correspondence{Bassel Al Omari at \email{balomari@meta.com}}

\begin{document}
\maketitle

\section{Introduction} 
\label{sec:introduction}
Language model agents have advanced rapidly in domains such as mathematics, coding, and computer use, where candidate actions can be evaluated accurately and efficiently: a mathematical answer can be checked against a reference, a program run against a test suite, and a computer-use task verified against its target state. These evaluation functions provide the reward signal that lets agents iterate and improve, driving rapid progress on benchmarks such as SWE-bench~\citep{jimenez2024swebench} and Terminal Bench~\citep{merrill2026terminalbench}.

Progress has been slower for AI research agents (AIRAs) that autonomously propose, implement, and evaluate their own experiments, despite recent efforts such as AIDE~\citep{jiang2025aide}, \airadojo~\citep{toledo2025ai}, \atlas~\citep{hambardzumyan2026aira2} and MARS~\citep{chen2026mars}. Frontier machine learning research lacks cheap feedback: proposing or modifying candidate code can be quick, but executing it to train a model and measuring its performance can consume hours to days or decades of GPU time. Considering an agent can propose far more candidates than it can afford to run, the primary lever on research progress becomes \emph{research preference}: deciding which research directions are promising enough to allocate compute budget to, and which directions to drop.

\begin{figure}[t]
    \centering
    \includegraphics[width=\linewidth]{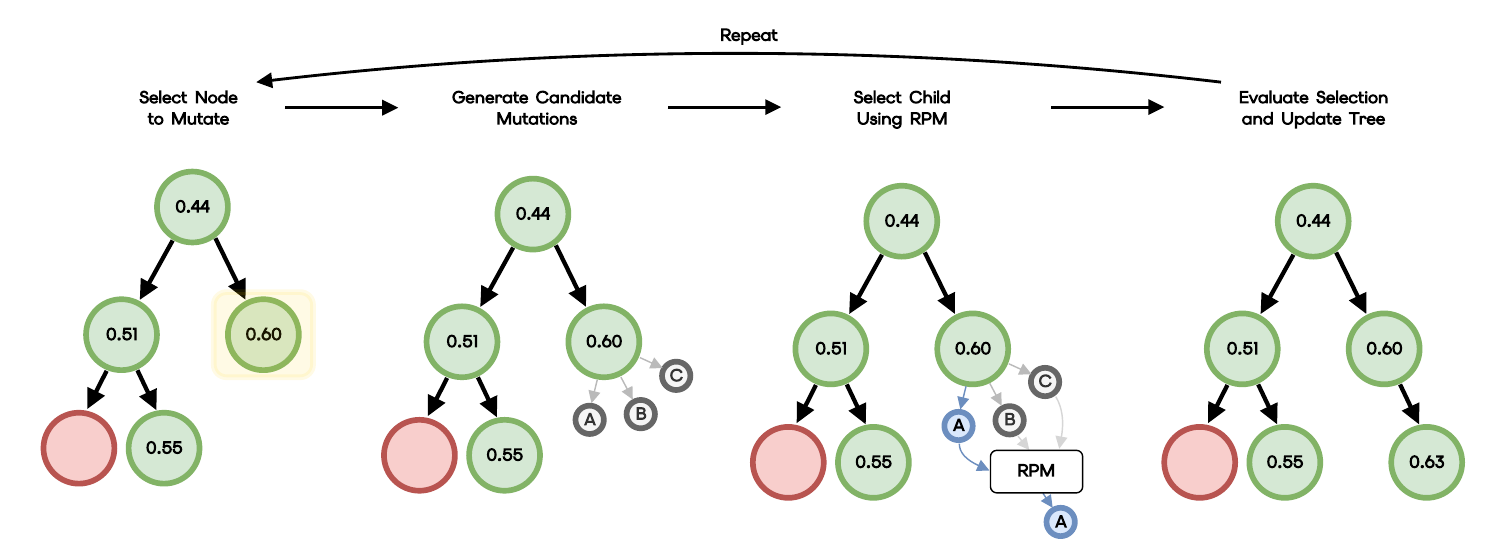}
    \caption{\textbf{Overview of an AI research agent with Research Preference Model (RPM) augmented child creation.} Each node represents a  solution within the search tree, labeled with its evaluation score. The agent (1) selects a promising parent node from the active tree, (2) generates a batch of candidate child solutions, (3) utilizes the RPM to select the most promising candidate, using the full context of previously executed solutions and (4) executes and scores only the selected candidate to expand the tree, to reduce the compute overhead of running unpromising candidates (\Cref{sec:aira_mlwm_integration}).}
    \label{fig:mlwm_augmented_airadojo}
\end{figure}

We address this allocation problem directly by introducing a dedicated \textbf{AI Research Preference Model (RPM)} into an AIRA. The RPM receives the full context of previously executed solutions, and uses it to decide which of multiple newly generated candidate solutions will be most valuable to execute next. 

We summarize our contributions below:
\begin{itemize}
    \item We introduce AI Research Preference Models, a framework for efficiently allocating compute to the most promising candidate solutions in AI research agents.
    \item We develop \textit{Inference-only RPMs}, frozen-weight LLMs that reason over candidate plans, code and previous solutions, and demonstrate that it raises performance of the \airadojo{} agent on \airsbench{} from $0.684$ to $0.711$.
    \item We further propose \textit{Agentic RPMs}, an extension of Inference-only RPMs with the ability to run small-scale pilot experiments. Integrated within \airadojo{}, Agentic RPMs further raise performance on \airsbench{} to $0.729$, approaching the validation oracle ceiling of $0.748$.
    \item On \airsbench{}, agents equipped with our best RPMs yield new state-of-the-art results on two tasks and match the unguided agent's 24-hour performance in roughly 15 hours, using less than two-thirds of its execution budget (\cref{fig:e2e_scores}).
\end{itemize}

\section{Background}

In line with the broader agent research literature, we view an agent as a computer system that is situated in some environment and is able to act autonomously in this environment in order to achieve its design objectives \citep{WoJe95}. In our setting, an AI research agent acts by generating and executing code. The objective is to produce an artifact (such as model weights, an optimised code snippet, or an answer to a question) that, when evaluated by some task-specific reward function, achieves a high score.

\subsection{AI Research Agent Benchmarks}
\label{sec:aira_benchmarks}

Recent benchmarks evaluate large language model agents across complex, long-horizon workflows. For software engineering, popular suites like SWE-bench \citep{jimenez2024swebench} assess an agent's ability to resolve real GitHub issues and modify multi-file codebases. Within the data science domain, benchmarks like MLE-bench \citep{chan2025mlebench} evaluate agent capabilities through structured machine learning engineering competitions.

We base our experiments on \airsbench{} \citep{lupidi2026airs}, a comprehensive suite of 20 machine learning tasks sourced from state-of-the-art papers. These tasks span diverse domains, including language modeling, mathematics, bioinformatics, and time series forecasting. \airsbench{} assesses agentic capabilities over the full research lifecycle (including idea generation, experiment analysis, and iterative refinement) without providing baseline code. Each task is rigorously specified by a problem, a dataset, a target metric, and a published state-of-the-art value. \iffalse \ag{sota seems to be mentioned a lot here. Just say a strong baseline or published baseline} \fi

Here the agents are placed in an environment with a training dataset, a set of test inputs, and 24 hours of access to an H200 GPU. The AIRA's goal is to write code that trains a model and produces a \texttt{submission.csv} with predictions on the test inputs. The agent may choose to run code that reports a validation score (e.g., from using cross-validation) that it can use for guiding search. The true test score (produced by comparing the \texttt{submission.csv} to the ground truth labels) is hidden from the agent. 

To aggregate performance across heterogeneous task metrics, \airsbench{} defines a Normalized Score for agent $a$ on task $t$:
\begin{equation}
    \text{NS}_{t}^a = \frac{\phi_t(s_{t}^a) - \phi_t(s^\mathrm{min}_t)}{\phi_t(s^\mathrm{sota}_t) - \phi_t(s^\mathrm{min}_t)},
    \label{eq:normscore}
\end{equation}
where $s_t^\mathrm{min}$ is the worst score observed across all agents, $s_t^\mathrm{sota}$ is the most recent public SOTA score as of the benchmark's publication, and $\phi_t$ is a non-linear log transform, defined as $\phi_t(s) = -\log_{10}(|s - s^\mathrm{opt}_t|)$ to properly weight exponential progress near optimal bounds ($s_t^\mathrm{opt}$). Under this metric, $\text{NS}=0$ corresponds to the minimum baseline and $\text{NS}=1$ to the public SOTA, while invalid or failed submissions receive a score of $0$. The Average Normalized Score reported on this benchmark averages the Normalized Score uniformly across all tasks and seeds.

\subsection{AI Research Agent Scaffolds}
\label{sec:scaffolds}

To systematically analyze and improve AI research agents, we decompose an AIRA into an LLM backbone and an algorithmic \textit{scaffold}. While the backbone provides core reasoning capabilities, the scaffold defines the decision logic and search strategy (ranging from simple linear loops to complex tree search) that govern how candidate solutions are generated, evaluated, and iteratively refined. A single scaffold like \airadojo{} \citep{toledo2025ai} or Claude Code \citep{anthropic2025claudecode} can be instantiated with different backbones, such as Claude Opus \citep{anthropic2026claude48} or GPT-5 \citep{singh2025openai}.

These candidate solutions can be viewed as nodes within an evolving solution graph, where each node stores concrete artifacts like code scripts, execution logs, and metric scores. The agent explores this graph by selecting a parent node and mutating its contents to produce a child node. Inspired by this evolutionary computation framework, we characterize the core design space of search scaffolds along three primary axes:
\begin{itemize}
    \item \textbf{Parent Selection:} The strategy used to identify which historical solutions, trajectories, or ideas are most promising to build upon next.
    \item \textbf{Child Creation:} The process of taking the selected parent solutions and prompting the LLM to generate a new candidate child solution. The localized prompts used to drive \emph{child creation} are referred to as \textit{operators}. 
  
    \item \textbf{Final Solution Selection:} The criteria used to evaluate the accumulated bank of candidate solutions and determine which single solution to return as the final output.
\end{itemize}

Existing AIRA architectures implement these mechanisms in fundamentally different ways. For example, \mlgym~\citep{nathani2025mlgym} operates as a linear search scaffold that heavily simplifies parent selection by always choosing the most recent node as the parent, relying on a single mutation operator to iteratively refine it. In contrast, we build upon \airadojo, an evolutionary tree-search framework that uses greedy parent selection to always mutate the node with the highest current validation score. To orchestrate child creation, \airadojo employs a specialized suite of mutation operators tailored to distinct engineering phases, specifically \draft, \improve, and \debug. Finally, for final solution selection, \airadojo submits the node that achieved the highest validation score across the entire search tree.

Ultimately, this loop produces an expanding bank of candidate solutions, each paired with its evaluation score. It is over this growing set that a preference model can intervene, deciding which candidates are worth the expense of execution. It also serves as a valuable dataset to train and evaluate an RPM.

\section{AI Research Preference Models}
\label{sec:machine_learning_world_models}

Frontier machine learning research lacks low-cost feedback: while proposing candidate solutions is fast, executing them to train a model can consume hours or days of GPU compute. Faced with this bottleneck, research progress heavily relies on predicting the value of pursuing candidate research directions.

To address this challenge, we introduce AI Research Preference Models (RPMs) to guide experimental allocation within research agents. Through initial experimentation, we observed that language models perform unreliably when forecasting absolute metrics or execution outcomes. Consequently, an RPM reformulates experimental allocation as a preference ranking problem: ranking candidate solutions to select the most promising paths before dedicating compute to pursuing them.

We explore RPMs leveraging varying ranges of test-time compute:
\begin{itemize}
    \item \textbf{Inference-only RPMs} (\Cref{sec:inference_only_mlwm}): Rank candidate solutions using lightweight reasoning over search history and code diffs.
    \item \textbf{Agentic RPMs} (\Cref{sec:agentic_mlwm}): Allocate additional compute to run rapid sandbox pilot experiments prior to ranking.
\end{itemize}

Designed as a scaffold-agnostic component, RPMs can interface with a wide variety of AIRA architectures. In this work, we investigate integrating RPMs into the \airadojo evolutionary tree-search scaffold (\Cref{sec:aira_mlwm_integration}). We target the child-creation phase, where creating a single child mutation is replaced with generating $N$ candidate modifications in parallel and using an RPM-guided tournament to select the most promising solution before committing GPU compute.

\subsection{Inference-only RPMs}
\label{sec:inference_only_mlwm}

The ``LLM-as-a-Judge'' paradigm \citep{zheng2023judging} demonstrates that LLMs can rank technical solutions with reasonable fidelity using internal intuition and reasoning.
Motivated by this approach, we experiment with purely querying pretrained LLMs as an inexpensive preference model to select between research ideas. To understand how visibility into the AIRA's search space affects the RPM's selection quality, we experiment with varying the count of previously explored solutions visible to the RPM and the count of suggestions it selects between.

To develop the chosen prompt even further, we leverage MIPROv2 from the DSPy framework \citep{opsahl2024optimizing}, a widely adopted baseline for robustly optimizing prompt instructions. The prompt optimizer generated an instruction set that directs the RPM to conduct a more structured analysis of each solution and remain tolerant of minor, fixable issues. Further details on the prompt are provided in Appendix~\ref{app:prompt_optimization}.

\subsection{Agentic RPMs}
\label{sec:agentic_mlwm}
\paragraph{\textbf{Motivation}} A core practice in software engineering, and machine learning research is rapid prototyping. To comprehensively understand and validate the potential or the feasibility of a novel idea, researchers tend to quickly run small-scale pilot experiments before launching a full-volume large-scale experiment. Inspired by how pilot experiments inform the possible outcome and assist decision making, we develop Agentic RPMs where an agent can use multiple predefined tools in a sandbox environment to conduct pilot experiments before making a decision.

\paragraph{\textbf{Agentic Workflow}}  Concretely, a pilot experiment is conducted via multi-turn interaction between the language model agent and the environment, interleaving chain-of-thought reasoning, tool calling, and receiving environment feedback~\citep{yao2023react}. The sandbox environment for the Agentic RPM is an exact clone of the environment for AIRA, including the access to a single H200 GPU. Meanwhile, the agentic RPM has access to the training dataset and unlabeled test dataset, together with necessary pre-installed Python packages in this environment, similar to the AI research agent. To interact with this environment, we provide the  agent with a set of tools: \texttt{python, bash}, and \texttt{submit\_solution}. The \texttt{python} tool and \texttt{bash} tool allow executing any Python code or Bash code, respectively, and then return execution results. With the outcome of the pilot experiment, the agent can summarize and submit the experimental findings with the tool \texttt{submit\_solution}. Notably, the agent is only required to submit the summary and analysis of the pilot experiment, but not to make a final selection among candidate solutions. The \texttt{submit\_solution} tool will return the remaining time budget. If the remaining time budget surpasses a specific threshold, the agent would be prompted to run further experiments to make the best use of the time budget and computation resources. Finally, the candidate solution, the task description, and all submitted pilot experiment findings are input to a language model to select the best solution. Full implementation details on agentic RPM are provided in Appendix \ref{app:agentic_mlwm_further_details}.

% \paragraph{\textbf{More Informative Experiments}} During our study, we find that the pilot-experiment agent can be too conservative in scheduling the time budget, leaving a large portion of the time budget unused at the first call of \texttt{submit\_solution}. Even though we could prompt the pilot-experiment agent to run more pilot experiments after the initial submission, the follow-up experiments are often limited to hyperparameter tuning of previous ones. Over conservative time budget scheduling and repetitive follow-up experiments jointly lead to less informative pilot experiments. To deal with this problem, we use two mechanisms to elicit more informative follow-up experiments. First, we \emph{overstate} the remaining time budget in the prompt (reporting it as several times larger than it truly is), discouraging the agent from stopping prematurely. Second, after each \texttt{submit\_solution} call, a separate feedback model reviews the findings so far and either proposes the single most informative next experiment or signals that the evidence is already sufficient, in which case we end the loop; its proposal and the remaining budget are then returned to the agent to guide the follow-up experiment. 

\subsection{AIRA Integration}
\label{sec:aira_mlwm_integration}
While an RPM can intercept multiple stages of an agent's search scaffold, we focus our implementation on augmenting the \textit{child-creation} phase, as illustrated in Figure~\ref{fig:mlwm_augmented_airadojo}. In an evolutionary AIRA scaffold, child creation fundamentally encompasses two sub-steps: child candidate creation and child candidate selection. By default, \airadojo{} generates a single candidate solution during candidate creation, which is automatically selected to be executed, evaluated, and added to the search tree. We modify this pipeline by first expanding child candidate creation: a chosen parent node is mutated by applying operators $N$ times independently in parallel to yield $N$ unexecuted child candidate solutions. During child candidate selection, we then introduce the RPM which evaluates these $N$ candidates alongside historical trajectory context, conducting pairwise comparisons in a tournament knockout structure to select the single candidate for full execution.

To ground each comparison in the search so far, we also provide the RPM selected context from the tree. Context nodes are pulled via a BFS traversal of the already-explored tree starting from the parent, collecting up to $K$ non-buggy nodes from earlier in the search; each of these previously-evaluated solutions is presented alongside the validation score it obtained.

\section{Experimental Setup}
\label{sec:experimental_setup}

\subsection{End-to-End Evaluation}
\label{sec:e2e_eval_setup}
We integrate RPMs into the child-creation phase of \airadojo, and evaluate this augmented scaffold against the 20 publicly released \airsbench{} tasks that fall under the text and tabular modalities.
Following the original \airsbench{} evaluation protocol, the \airadojo evaluations are provided 24 hours of access to a single H200 per task and are repeated for 10 seeds.
To avoid the generalization gap which undermined long-horizon search in \airadojo, we integrate the Hidden Consistent Evaluation protocol from \atlas. Both the \airadojo operator used to generate the child candidates and the RPMs share a Qwen3.6-27B \citep{qwen3.6-27b} backbone, standardizing our setup on a high-performing open-weights model for code reasoning. Maintaining an identical model across child creation and selection ensures that all observed improvements are driven by the framework rather than a stronger selection backbone.

To contextualize these results, we first compare against the default No-RPM baseline. In this setup, child selection defaults to uniform random selection among generated candidates, which corresponds to vanilla \airadojo{} in expectation. We also compare against a Test Oracle and a Validation Oracle, that are constructed by executing all candidates at each step and choosing the highest scorer. Only the compute time of the selected candidate counts towards the 24-hour limit.
While neither is viable online (the Test Oracle utilizes privileged test-set information, and the Validation Oracle requires a prohibitive compute overhead to execute every candidate), they serve as ceilings for greedy child selection.

\subsection{Offline Evaluation}
\label{sec:offline_eval}
The full end-to-end evaluations detailed in \Cref{sec:e2e_eval_setup} are computationally expensive, requiring 200 H200 GPUs for 24 hours. To enable quick iterations and guide development of our RPMs before running full end-to-end evaluations, we produce an offline evaluation dataset compiled from previous \airadojo runs on a separate set of 40 unreleased \airsbench{} tasks from the image, video and audio modalities. The development and evaluation sets are split by modality to avoid task contamination.
The previous \airadojo runs followed the standard \airsbench{} evaluation protocol of dedicating 24 compute hours with a single H200 and evaluating 10 different seeds. The previous runs were conducted with gpt-oss-120b~\citep{openai2025gptoss}, GPT-4o~\citep{openai2024gpt4osystemcard} and CWM~\citep{copet2025cwm} as the LLM backbone.
From these runs, we extract 1,000 sibling node pairs, including their plans, code, and search tree history. In selecting these pairs, we discard near-ties with a normalized test-metric gap below 0.01, to prevent negligible, run-to-run metric noise from confounding the evaluation signal. The RPM is evaluated on its accuracy in selecting the node with the highest test score in its subtree. We choose this ground-truth label to ensure the RPMs look past immediate performance, explicitly rewarding candidates with strong long-term fixability and extensibility. We acknowledge that this label inherits a bias from the original greedy search policy, where nodes with stronger early scores are favored during search, expanding their subtrees and giving them greater opportunity to reach high scores. Random selection establishes a 50\% baseline accuracy floor. In these offline evaluations, GPT-5 \citep{singh2025openai} is used as backbone for the RPMs.

\section{Results}
\label{sec:results}

\subsection{End-to-End Evaluations}
\label{sec:e2e_eval}

\begin{figure}[t!]
    \centering
    \includegraphics[width=0.49\linewidth]{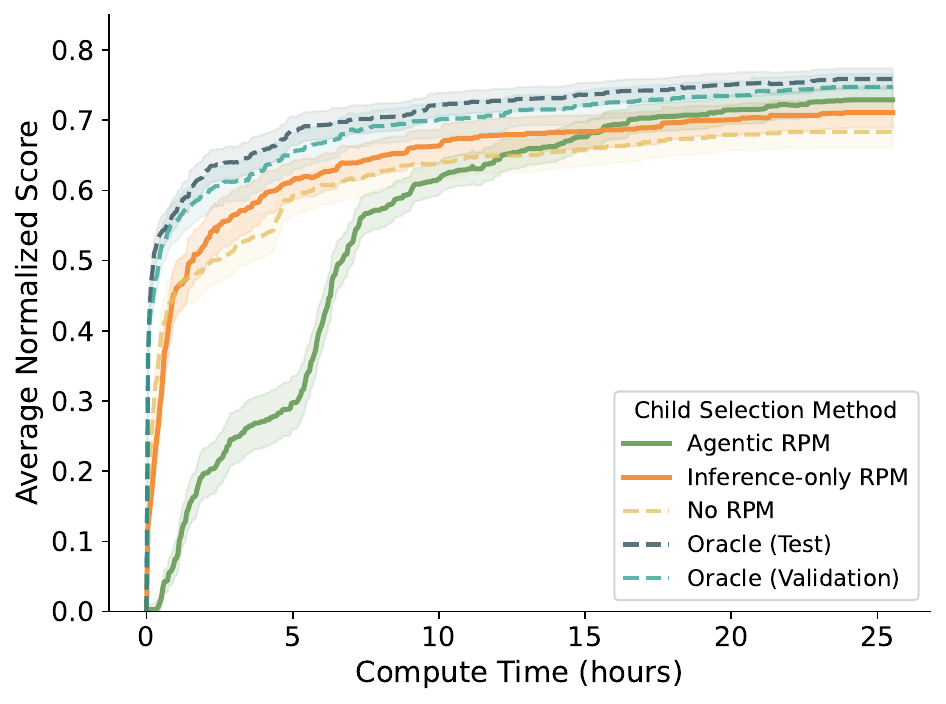}
    \hfill
    \includegraphics[width=0.49\linewidth]{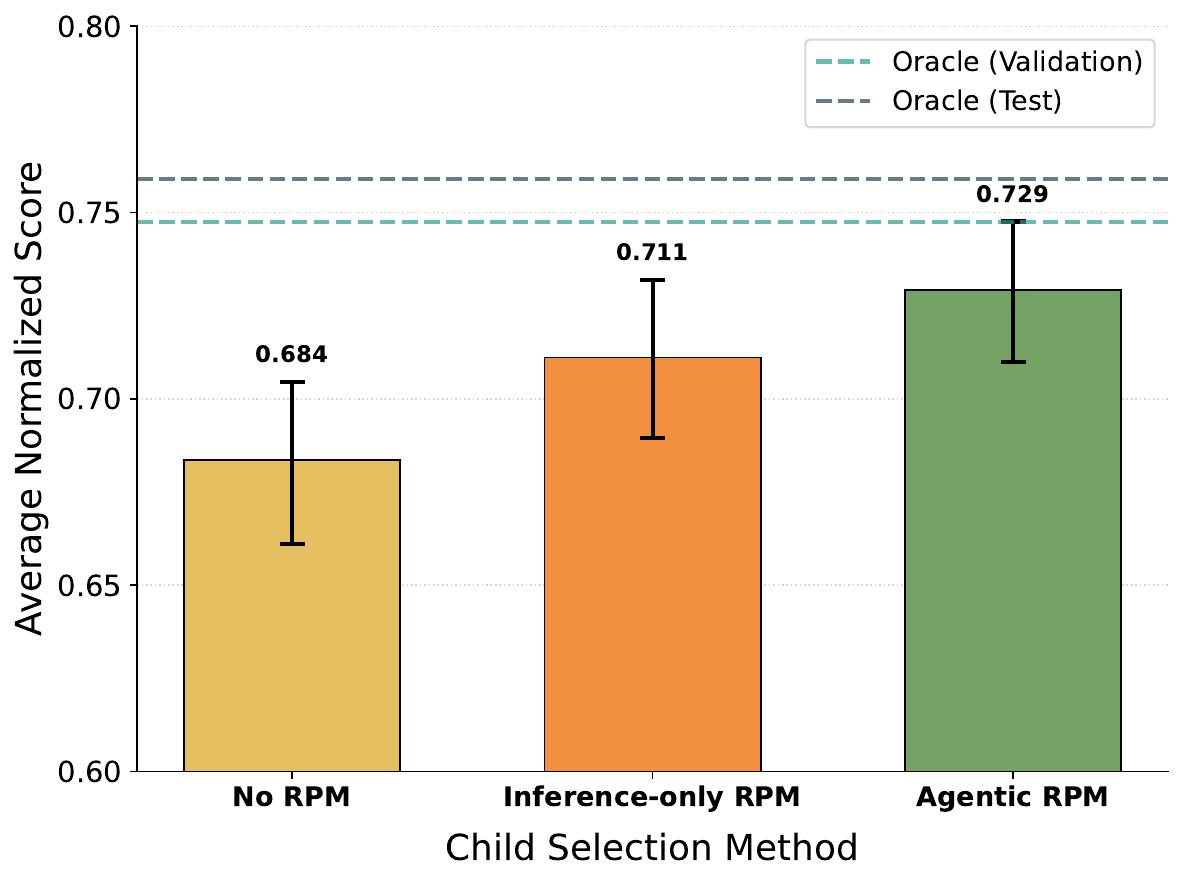}
    \caption{\textbf{Evaluation of RPM-augmented \airadojo.} Average normalized scores over time (left) and final performance (right) for \airadojo with RPM-augmented child selection on \airsbench, with error bars indicating the 95\% confidence intervals. Dashed lines indicate the No-RPM baseline and immediate validation or test oracles. The Inference-only RPM (orange) yields steady gains over the baseline (yellow). While per-step proxy experiments slow its early trajectory, the Agentic RPM (green) leverages this compute overhead to surpass other methods.}
    % \ag{Random idea - but given the SOTA section later on, what happens if you plot the max score attained by a seed rather than the average over seeds - just a random possible interesting demonstration?}
    % Average normalized scores for \airadojo with RPM-augmented child selection on \airsbench~over time (left) and final performance (right). Dashed lines denote the theoretical upper bounds of perfect selection (Oracles) according to immediate test or validation score, alongside the standard \airadojo baseline with no RPM. Inference-only RPM (orange) provides steady gains over the baseline (yellow). In contrast, Agentic RPM (green) runs small-scale proxy experiments at every step. This heavy per-step compute consumption slows its early progress along the time axis, but ultimately pushes performance near the validation oracle. Shaded regions (left) and error bars (right) denote bootstrapped confidence intervals~\citep{agarwal2021deep}.
    \label{fig:e2e_scores}
\end{figure}

We present the end-to-end performance of integrating our RPMs within \airadojo, in Figure~\ref{fig:e2e_scores}.
In these evaluations, \airadojo{} is configured to generate 15 child candidate suggestions at each operator step, leveraging our offline finding that expanding the candidate pool size systematically improves selection performance (Section~\ref{sec:inference_scaling}). The Inference-only RPM evaluates these pairs using the ``LLM-as-a-judge'' reasoning prompt presented in Section~\ref{sec:inference_only_mlwm}, including the scores of historical nodes from the search tree. For the Agentic RPM detailed in Section~\ref{sec:agentic_mlwm}, to balance overall compute budgets, we only deploy this selection mechanism for the \draft and \improve operators, reverting to random selection during \debug steps.

As shown in Figure~\ref{fig:e2e_scores}, our methods navigate the trade-off between decision quality and compute time in fundamentally different ways. Incurring no candidate-execution cost, the Inference-only RPM (orange) provides immediate and steady gains over the baseline with no RPM (yellow). Conversely, the Agentic RPM (green) exhibits a slow rise early in the run. Considering it runs small-scale proxy experiments in a sandbox at every step, its heavy per-step compute consumption slows early progress along the time axis. However, this rigorous per-step evaluation eventually triggers a sharp performance acceleration, ultimately matching or exceeding both the Inference-only RPM and the No-RPM baseline. Ultimately, both variants beat the unguided baseline's final score of 0.684, with Inference-only reaching 0.711 and Agentic reaching 0.729, narrowing the gap toward the validation-oracle (0.748) and test-oracle (0.759) ceilings.

\paragraph{\textbf{Significance Testing}} To evaluate statistical significance, we report the \emph{probability of improvement}, defined as the likelihood that a randomly sampled run of one method outperforms another on a randomly selected task. We compute task-stratified bootstrap distributions using \texttt{rliable} \citep{agarwal2021deep}, where a probability of $0.5$ denotes no difference. The results show that the Inference-only RPM and the Agentic RPM achieve a statistically significant edge over the default \airadojo baseline (No RPM), yielding average improvement probabilities of 0.5923 and 0.5913, respectively, with the 95\% confidence intervals lower bounded at 0.5066 and 0.5018, strictly excluding the 0.5 mark of random chance.

\paragraph{\textbf{State-of-the-Art Breakthroughs}} These guided search capabilities translate directly to new state-of-the-art (SOTA) milestones on established benchmarks, to the best of our knowledge. On WinoGrande \citep{sakaguchi2021winogrande}, \airadojo with the Agentic RPM achieves an accuracy of 94.1\%, comfortably surpassing the previous agentic SOTA of 90.4\% reported by \citet{hambardzumyan2026aira2}. During this run, the agent fine-tunes a Qwen2.5-14B-Instruct model \citep{qwen2.5} via LoRA on data with shuffled labels, then averages prediction logits across original and shuffled label orderings at inference to eliminate position bias. Similarly, on SVAMP \citep{patel2021nlp}, \airadojo with the Inference-only RPM reaches 95.7\% accuracy, eclipsing the prior human SOTA of 94.2\% from \citet{zhong2026achieving}. Here, the agent designs few-shot prompts that instruct the model to cleanly isolate relevant numerical data from distracting context, then generates ten independent reasoning paths by sampling Qwen2.5-7B-Instruct with increased temperature, and resolves the final prediction using a majority vote.

\paragraph{\textbf{Research Efficiency}} Beyond absolute performance gains, both preference models significantly accelerate search velocity. While the unaugmented No-RPM baseline requires the full 24-hour allocation to reach its final score of 0.684, our RPM-guided approaches reach this identical performance threshold significantly faster. The Inference-only RPM matches this baseline score in 14.88 hours (a $1.61\times$ speedup), while the Agentic RPM achieves it in 15.50 hours (a $1.55\times$ speedup). This allows both methods to match standard performance while using approximately $1.5\times$ less compute budget.

\paragraph{\textbf{Impact of Selection Quality}} To confirm how decision quality drives performance, we retrospectively analyze the selection advantage of each RPM throughout the runs. Selection advantage measures the average difference between the chosen candidate's score and the overall batch mean. As expected, random selection yields a selection advantage of roughly $0.0$ in expectation, whereas the Inference-only RPM achieves significantly higher selection quality, and the Agentic RPM yields the highest advantage on average. Crucially, we find a strong positive correlation between selection advantage and final normalized score (Pearson $r=0.55$, Spearman $\rho=0.56$), confirming that more accurate candidate selection translates to better end-to-end AIRA performance. Further details are provided in Appendix \ref{app:selection_quality_e2e_impact}.

\subsection{Offline Evaluation and Further Analysis}
\label{sec:offline_eval_results}

To guide development of our RPMs before running full end-to-end evaluations, we leverage the offline framework from Section~\ref{sec:experimental_setup}. Full 24-hour online runs are too computationally expensive for quick iteration, and thus this offline setup allows us to tune our models, analyze their core behaviors, and derive clear trends.

We observe that the best configuration of Inference-only RPMs is surpassed in predictive accuracy by our Agentic RPMs, with both exceeding the random baseline, matching the trend ultimately observed in our end-to-end evaluations. 

\subsubsection{Inference-only RPM}

\paragraph{\textbf{Inference Scaling}} 
\label{sec:inference_scaling}

\begin{figure}[t!]
    \centering
    \includegraphics[width=\linewidth]{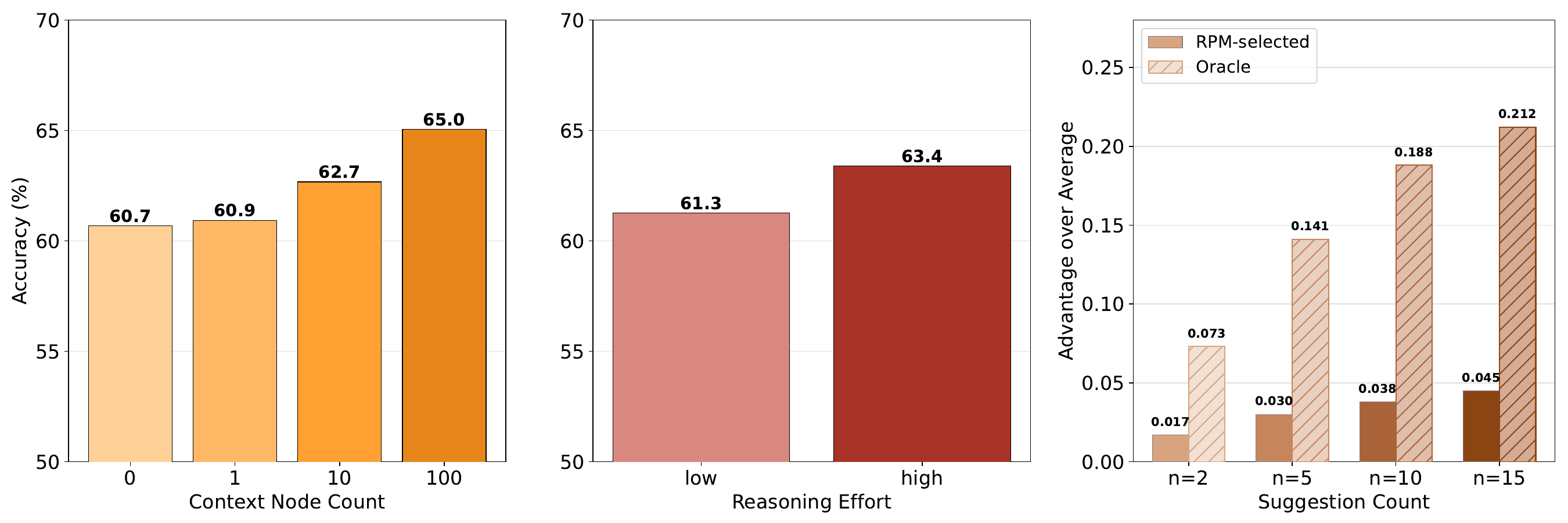}
    \caption{\textbf{Offline evaluation of Inference-only RPM scaling properties.}
    Increasing search-tree context provided gives improved predictive accuracy (left). Allocating a higher reasoning budget yields better selection (middle). Expanding the candidate suggestion pool reliably improves both the oracle-selection limit and the RPM's selection performance (right), measured by the average difference between the selected solution's score and the average score of all candidates.  
    }
    \label{fig:inference_scaling}
\end{figure}

We structure our offline analysis around three key dimensions that dictate the RPM's visibility and analytical capacity: \emph{context size} (the number of historical code solutions and validation scores provided from the tree), \emph{suggestion count} (the number of candidate child solutions), and \emph{reasoning budget} provided to the LLM.

Our offline tests reveal three clear scaling behaviors. First, \emph{context scaling} demonstrates that supplying the RPM with a deeper history of historical search tree nodes and their validation scores consistently improves child-selection judgment. Second, \emph{suggestion scaling} demonstrates that expanding the candidate pool systematically increases the selection advantage, measured as the average difference between the selected candidate's score and the overall batch mean. The oracle's selection advantage rises significantly with pool size, a trend our RPM successfully captures to extract higher-quality solutions from larger candidate batches. Third, \emph{reasoning scaling} reveals that increasing the reasoning budget allocated to the LLM judge steadily improves its selection accuracy. 

These results motivated using a large suggestion count (15 suggestions) and to maximize the context nodes provided (by providing the maximum amount that can fit within the LLM's context window), and high reasoning budget parameters for the RPM in our end-to-end evaluation.

\paragraph{\textbf{Ensembling}} 
We evaluate three frontier models, GPT-5 \citep{singh2025openai}, Claude Opus 4.8 \citep{anthropic2026claude48} and Gemini 3.1 Pro \citep{google2026gemini31pro}, individually and via two aggregation techniques: a mechanical majority vote and an LLM-Arbiter ensemble that ingests the reasoning traces of all three models before making a final decision. Individual models' performances range between 64.66\% to 67.44\% accuracy. Aggregating their diverse reasoning traces further mitigates errors where majority vote increases accuracy to 68.04\%, while the LLM-Arbiter ensemble achieves the highest overall offline accuracy of 69.35\%. A full breakdown of these configurations and their corresponding results is compiled in Appendix~\ref{app:appendix_ensembling}.

\paragraph{\textbf{Reasoning Analysis}} To get a better understanding of the RPM's decision-making, we analyze its generated reasoning traces. We find that referencing prior evidence, correctness of implementation or the pretrained backbone in the justification yields higher selection accuracy compared to when these are omitted. We also find, that citing more unique values from historical context also yields improved selection accuracy. Further details on the reasoning analysis are provided in Appendix \ref{app:reasoning_analysis}.

\subsubsection{Agentic RPM}
\label{sec:agentic_rpm_offline_eval_results}

\begin{figure}[t!]
    \centering
    \begin{minipage}{0.55\textwidth}
    \includegraphics[width=\linewidth]{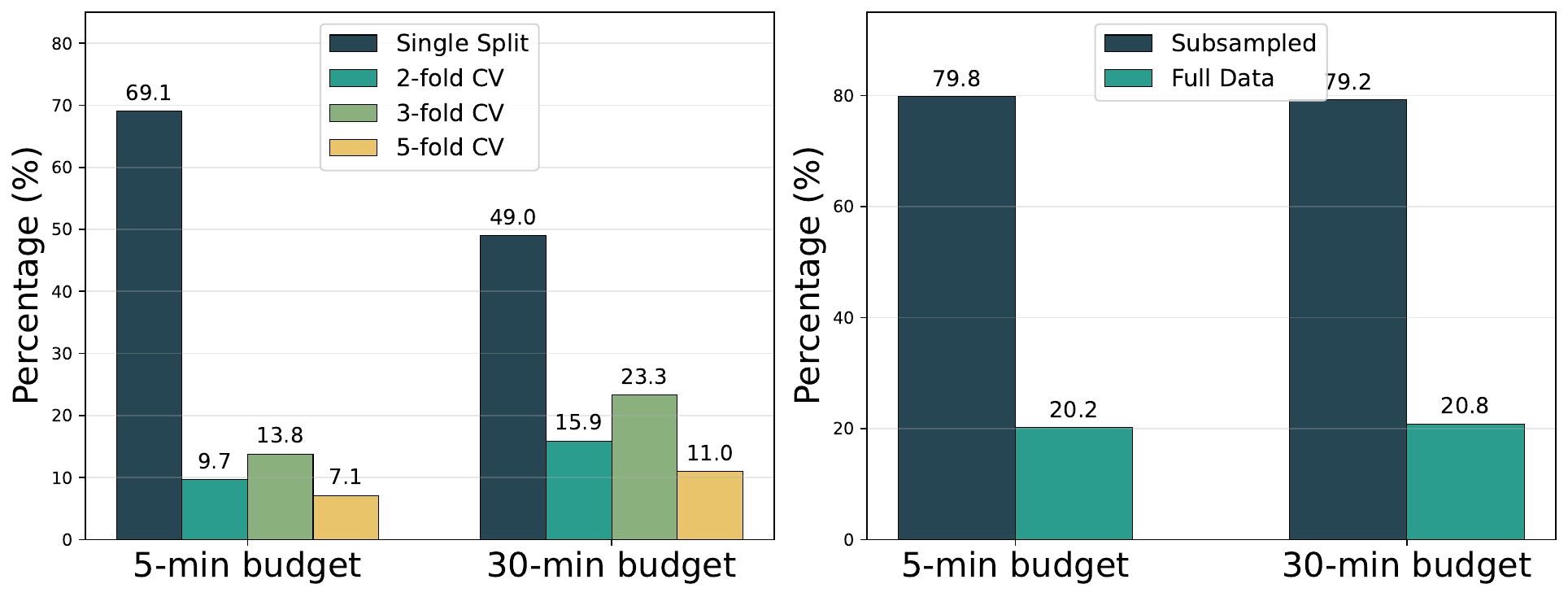}
    \end{minipage}
    \begin{minipage}{0.42\textwidth}
    \includegraphics[width=\linewidth]{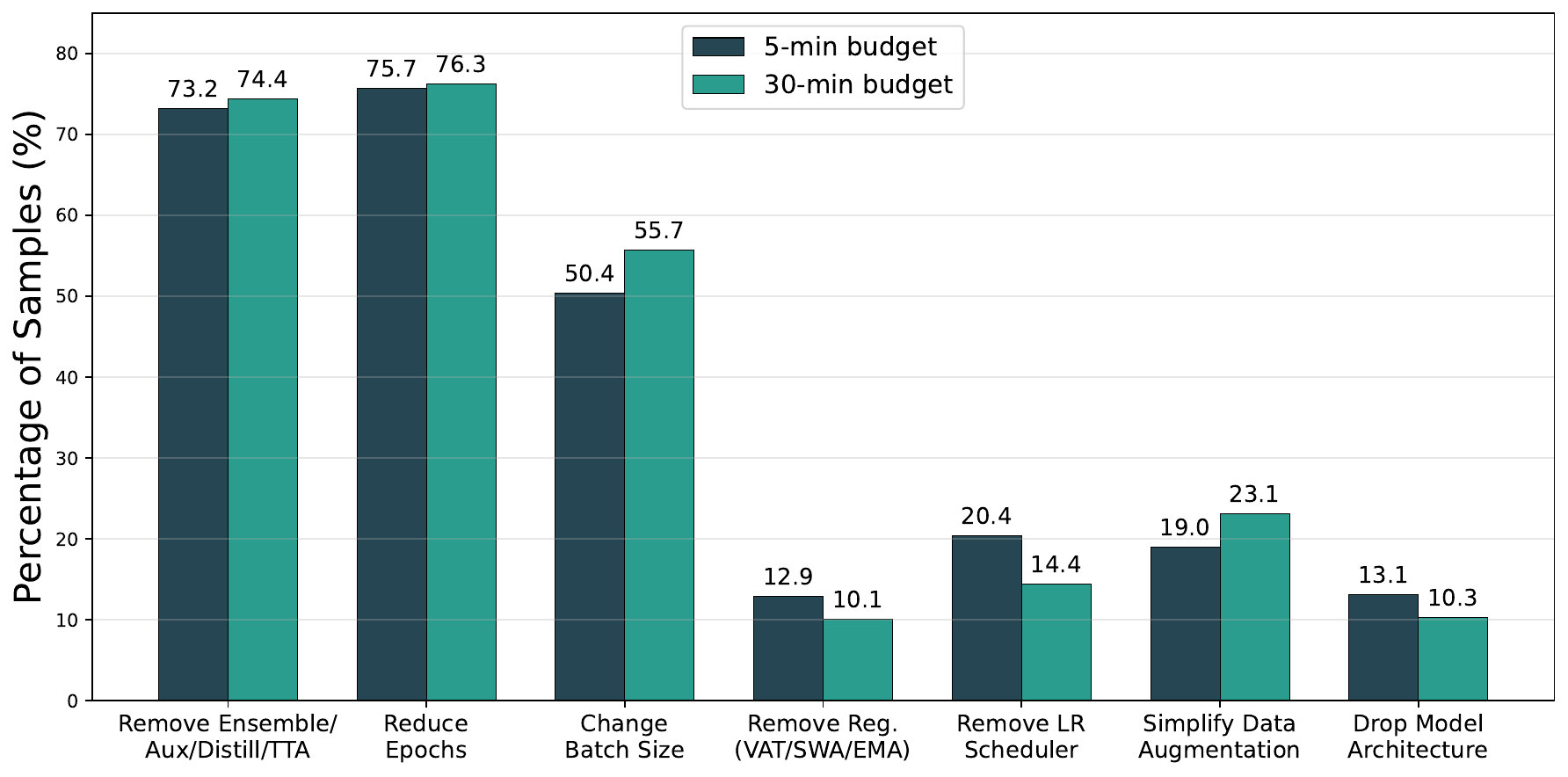}
    \end{minipage}
    \caption{\textbf{What strategy the Agentic RPM uses.} The Agentic RPM frequently runs simplified versions of the original candidates. Agents save time by running reduced cross-validation split counts (left) and sub-sampling the training data (middle). Both agents employ similar training adaptations, such as removing ensembling, reducing epochs and lowering the batch size.}
    \label{fig:arq_strategy_data_training}
\end{figure}

\begin{figure}[t!]
    \centering
    \begin{minipage}{0.22\textwidth}
    \vspace{-0.44cm}
    \includegraphics[width=\linewidth]{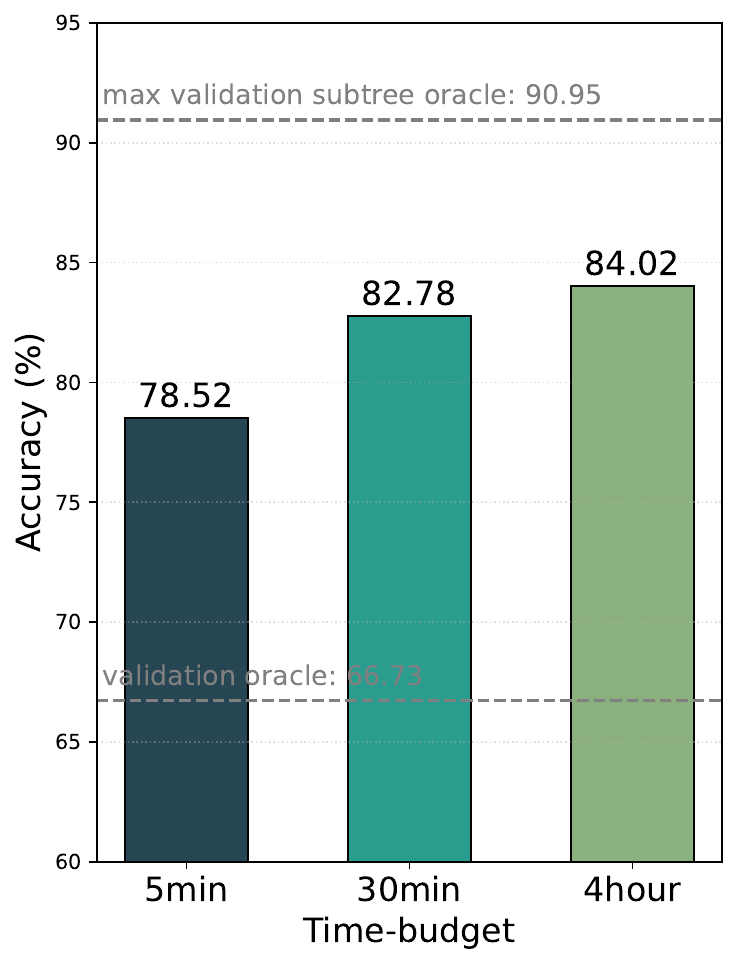}
    \end{minipage}
    \begin{minipage}{0.77\textwidth}
    \includegraphics[width=1.0\linewidth]{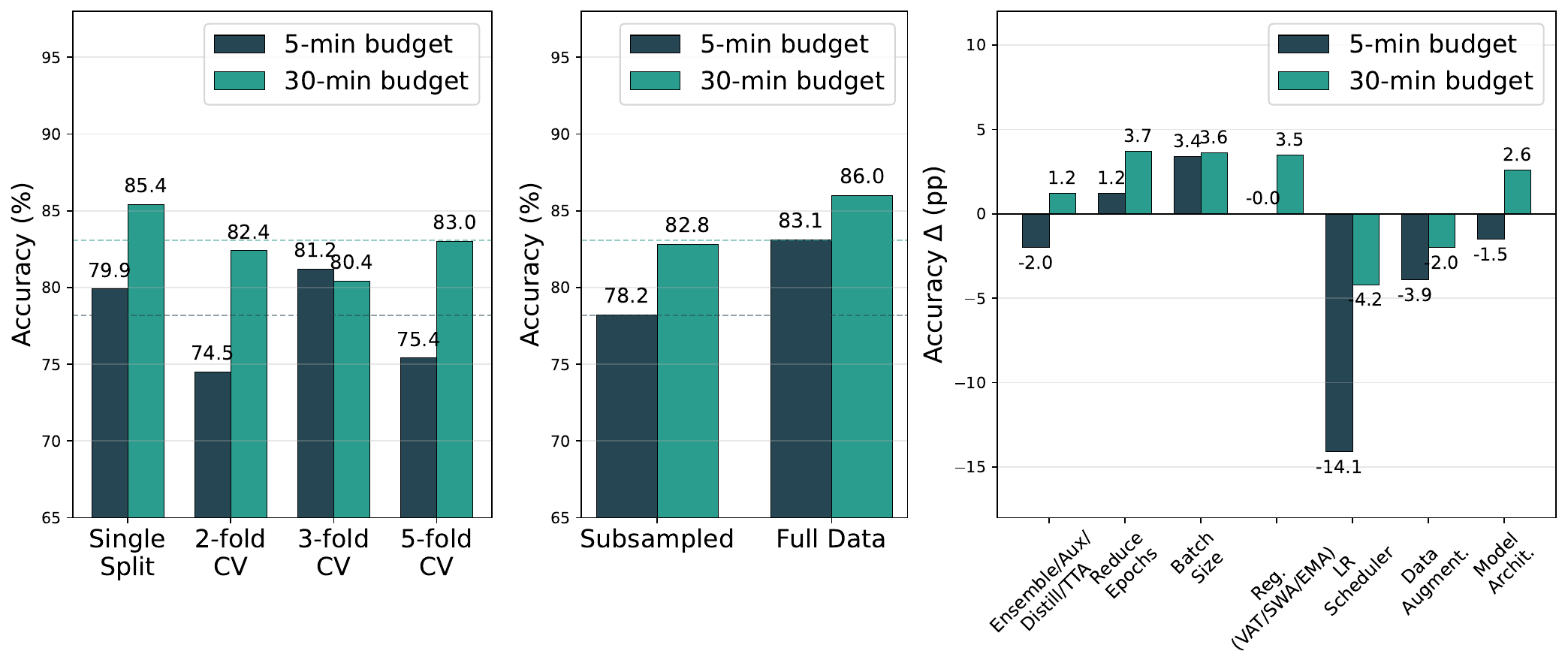}
    \end{minipage}
    \vspace{-1em}
    \caption{\textbf{How the Agentic RPM's strategy affects performance.} Increasing the time budget leads to better performance, but with diminishing returns (leftmost). Using more cross-fold validation splits does not correlate with better selection accuracy (middle left). Using full data brings an advantage over using sub-sampled data (middle right). Most training adaptations improve selection accuracy, except for the removal of the learning rate scheduler and simplification of the data augmentation approach (rightmost).}
    \label{fig:arq_strategy_accuracy}
\end{figure}

\paragraph{\textbf{Compute Scaling}}
We analyze how the agentic RPM scales with execution time limits on our offline benchmark. Increasing the available compute budget consistently improves selection accuracy, scaling from 78.52\% under a 5-minute constraint, to 82.78\% at 30 minutes, and peaking at 84.02\% with a 4-hour allocation. This scaling behavior demonstrates that allowing the agent more time to run, observe, and debug directly translates to higher-fidelity RPM estimations. Notably, the marginal returns obtained from increasing the time budget are relatively minimal considering the pilot-experiment agent shares the same time budget with the AI research agent and a 4-hour run of the Agentic RPM is prohibitively expensive. Therefore, we keep the time budget for end-to-end evaluation at 5 minutes.

\paragraph{\textbf{Proxy Strategy Analysis}}
To understand how agents identify promising candidates under strict time constraints, we use regular expression keyword matching to analyze the agent-generated Python code. Because full execution is impossible within 5- or 30-minute limits, agents actively simplify candidate code across validation, data sampling, and training. As shown in Figure~\ref{fig:arq_strategy_data_training}, both budgets favor single-split validation, data subsampling, and training adaptations (e.g., removing ensembles, reducing epochs, and lowering batch sizes), though the 30-minute agent utilizes multi-fold cross-validation more frequently. Notably, we provide some hints on possible proxy strategies in the prompt to the pilot experiment, as shown in Appendix \Cref{fig:arq_prompt} so the strategies are not entirely proposed by the pilot-experiment itself.

Our behavioral analysis in Figure~\ref{fig:arq_strategy_accuracy} reveals three key insights: (1) multi-fold cross-validation does not consistently improve accuracy due to frequent execution timeouts; (2) evaluating on full datasets provides significantly more reliable quality estimations than using subsampled data; and (3) removing learning rate schedulers or data augmentations triggers severe performance drops, as the underlying \airadojo tasks often rely on these exact training-level optimizations to succeed.

\section{Related Work}

\textbf{AI research agents and automated ML research.}
Our work sits within the literature that formalizes machine learning research as an agentic search problem. Early literature for automated scientific discovery and research assistance uses foundation models to generate ideas, write code, run experiments, analyze results, and draft papers
\citep{lu2024ai, aiscientistv2, schmidgall2025agentlab,
baek2025researchagent, ren2025survey, zheng2025automation}. Current work introduces AI research agents as search policies over a tree of candidate ML solutions, where each node corresponds to an executed or proposed solution and edges correspond to mutations, refinements, or other improvement operators \citep{toledo2025ai, hambardzumyan2026aira2}. This formulation directly motivates our work: RPMs target the core decision problem induced by such search trees, namely predicting which candidate solution should be executed before committing expensive compute. Other recent works improve the agent scaffold by adding modular search, targeted refinement, ideation agents, multi-agent specialization, or improved code interfaces \citep{mlestar, chen2026mars,
zhang2026ideate, li2024autokaggle, yang2024sweagent}. While these works improve agents, scaffolds, or benchmarks, none of them address the dominant cost in this setting, namely \emph{executing} the candidate solutions that the search proposes.

\textbf{Model-based methods for expensive search.}
In reinforcement learning, world models learn environment dynamics and can be used for planning \citep{schrittwieser2020mastering} or policy learning from model-generated rollouts \citep{sutton1991dyna, ha2018recurrent, hafner2023mastering}. At the \textit{meta}-level, related methods include Bayesian optimization \cite {snoek2012practical}, early stopping and model-guided search \citep{li2018hyperband, falkner2018bohb}, population-based online selection \citep{jaderberg2017population}, learned hyperparameter optimization \citep{chen2022optformer}, and learned optimization \citep{andrychowicz2016learning}. \citet{metz2022velotrainingversatilelearned} favor task configurations predicted to expedite optimizer meta-training, \citet{goldie2025how} use a distillation objective as opposed to online evaluation due to the time-cost of training models with every proposed algorithm, and \citet{wolf2026modelbased} present model-based meta-learning for algorithm discovery. These works share our goal of reducing expensive evaluation, yet none use preference ranking based on solution code, reasoning, and partial results to decide which proposed algorithm to evaluate next.

\textbf{Preference, judge, and reward models.} At its core, the RPM is a preference model over candidate solutions. Reward models trained from human preferences are central to reinforcement learning from human feedback \citep{christiano2017deep, stiennon2020learning, ouyang2022training} and their objectives connect to paired-comparison and ranking formulations \citep{bradley1952rank, liu2009learning}. Related preference supervision can also be constructed from expert trajectories without reward-model training or online environment rollouts \citep{chen2026agenticdpo}. Additionally, strong LLM judges can approximate human preferences on open-ended responses, subject to known biases \citep{zheng2023judging}. Recent work derives continuous scores from scoring-token logits to reduce ties and rank candidates without additional verifier training \citep{kwok2026verifier}, while \citet{qwen2026verification} argue that fixed coding-agent rewards can saturate as generators improve. These methods mostly evaluate completed responses, trajectories, or agent outputs, whereas the RPM ranks candidate solutions before expensive full training and evaluation.

Closest to our setting, \citet{zheng2026predictbeforeexecuting} also predict a pairwise preference between unexecuted ML solutions, but condition on a separately prepared, static data report and the two candidate code snippets, rather than on nodes from the surrounding search tree; the RPM instead grounds each comparison in the live search tree and the validation scores of solutions already executed. \citet{goldie2026discogen} propose training a judge or reward model to select promising leaves in tree-search research agents, but leave it unimplemented. The RPM realizes that proposal without model training, and extends it: where agentic verifiers probe code or user interfaces \citep{qwen2026verification}, our agentic variant runs small-scale \emph{pilot ML experiments}, partial training and evaluation runs, and chooses based on their measured results.

\section{Limitations}
\label{sec:limiations}
\begin{itemize}
    \item While our evaluations operate under the assumption that LLM inference calls will incur negligible cost in the future, in practice, current LLM inference incurs real-time latency. For the Inference-only RPM, which uses a self-hosted Qwen3.6-27B, inference latency totals 0.660 hours per 24-hour end-to-end run. Adjusting for this time budget yields a normalized score of 0.708 at 23.34 hours, a negligible drop from 0.711. Nevertheless, exact latency and monetary overhead vary by hosting infrastructure and LLM size.
    \item As the offline data come from prior greedy \airadojo{} runs (with different LLM backbones and, by design, different task modalities than the online setting), they are off-policy and biased relative to the online target, including the subtree-max label bias noted in \Cref{sec:experimental_setup}, and thus our main claims rest on the end-to-end results.
    \item We limit our RPM integration to the \emph{child creation} stage only. We also report initial results integrating RPMs within \emph{final-node selection} in Appendix \ref{app:pickyourpoison}. However, we do not observe significant improvement over validation-based selection, given the Hidden Consistent Evaluation protocol's strong test-validation generalization~\citep{hambardzumyan2026aira2}. We leave RPM integration within \emph{parent-selection} to future work.
    \item We describe the RPM as scaffold-agnostic because it inspects no scaffold-internal state, but we demonstrate it only in \airadojo{}'s child-selection step, with a single backbone (Qwen3.6-27B) on a single benchmark. The agentic variant additionally needs a sandboxed clone of the execution environment in which to run pilot experiments. We see no reason the approach would not transfer, but portability to other scaffolds and backbones is part of our future work.
\end{itemize}

\section{Conclusion}

An AI research agent can propose a candidate solution far faster than it can evaluate it. We introduced the \textbf{AI Research Preference Model (RPM)}, which ranks unexecuted candidates so that the agent can direct its execution budget toward promising candidates. This formulation avoids requiring the model to forecast absolute outcomes: it identifies the most promising candidate without taking on the challenging task of predicting what any candidate would score.

Our results show that RPM-guided candidate selection improves both final performance and research efficiency while keeping the underlying agent backbone and search operators fixed. More broadly, they show that candidate selection is a useful target for test-time compute: agents can invest computation not only in generating candidates, but also in deciding which candidates are worth executing.

AI research often lies on the less favorable side of the asymmetry of verification: candidate solutions are easy to generate but hard to verify through full training and evaluation \citep{wei2025asymmetry}. RPMs are designed to navigate this asymmetry. While they cannot reduce the cost of any single verification, they can reallocate it, spending the same budget on preferred candidates to reach better solutions sooner. By making this allocation part of the research loop, we hope this work encourages research agents that choose which solutions to run as carefully as they design them.

\clearpage
\bibliographystyle{assets/plainnat}
\bibliography{paper}

\clearpage
\beginappendix
% from Bhavul, ThomasMann
\lstdefinestyle{prompt}{
  basicstyle=\ttfamily\scriptsize,
  breaklines=true,
  columns=fullflexible,
  frame=single,
  backgroundcolor=\color{black!3},
  keepspaces=true,
  showstringspaces=false,
}

% Placeholder box for figures whose source files are not yet in figures/.
\newcommand{\missingfig}[2][0.7\linewidth]{%
  \fbox{\begin{minipage}[c][0.28\linewidth][c]{#1}%
    \centering\itshape Figure placeholder\\[4pt]%
    \ttfamily\footnotesize #2\\[4pt]%
    \normalfont\itshape(figure source not yet available)%
  \end{minipage}}%
}

\section{Inference-Only RPM}
\label{app:inference_only_rpm}
\subsection{Prompt Optimization}
\label{app:prompt_optimization}

We optimize the Inference-only RPM's prompt using Automatic Prompt Optimization with no underlying weight updates.

\subsubsection{Optimization Configuration}
We base our prompt optimization on MIPROv2 from the DSPy framework \citep{opsahl2024optimizing}. Starting from the handwritten baseline template presented in \Cref{fig:unoptimized_inference_only_prompt}, the meta-proposer generates 10 candidate prompt variations. The search space is explored via Thompson sampling over per-candidate $\text{Beta}(1,1)$ accuracy posteriors across 40 Bayesian minibatch trials. Underperforming templates are progressively pruned, while top-performing survivors undergo deeper validation runs to mitigate optimization-to-holdout shrinkage. We use GPT-5 as both the meta-proposer and the Inference-only RPM. We evaluate against an offline dataset of 990 samples, drawn similarly to the dataset presented in \Cref{sec:offline_eval} but without discarding near-ties.

\subsubsection{Results}

The optimization routine converges on a \emph{Principal Investigator} persona structured around a five-criterion evaluation rubric. This optimized prompt introduces three main shifts from the initial prompt:
\begin{itemize}
    \item \textbf{Structural Evaluation:} Moves from a flat list of decision rules to a structured walkthrough forcing the model to sequentially score problem-model fit and extensibility.
    \item \textbf{Shifting Bug Tolerance:} Replaces strict penalization of code bugs with an evaluation of core ideas, accepting code bugs if the candidate provides a more promising performance ceiling.
    \item \textbf{Strategic Context Utility:} Sharpens context node usage into an explicit mandate to discover open search gaps and actively penalize redundant directions.
\end{itemize}

The optimized prompt improves selection accuracy from $57.7\%$ to $59.0\%$.

\subsubsection{Prompt Templates}

We present the complete prompt templates for the Inference-only RPM before and after the automated prompt optimization procedure in Figures~\ref{fig:unoptimized_inference_only_prompt} and \ref{fig:optimized_inference_only_prompt} respectively.

\begin{tcolorbox}[colback=md-bg, colframe=md-bg, boxrule=0pt, left=6pt, right=6pt, top=6pt, bottom=6pt, enhanced, sharp corners, breakable]
\begin{lstlisting}[basicstyle=\ttfamily\small, breaklines=true, breakatwhitespace=true, columns=fullflexible, keepspaces=true, showstringspaces=false]
You are a strict judge selecting between TWO candidate solutions to the SAME machine learning task.

Your goal is to choose the candidate whose direction of exploration is more likely to eventually lead to a better long-term best test score, even if further refinements are needed. Focus on long-term potential rather than immediate performance.

Task description:
```markdown
{task_desc}

Context from various solutions to the same machine learning task. These are NOT the candidates you are judging.
{context_text}

Candidate A -- Plan:
{plan_A}
Candidate A -- Code:
{code_A}

Candidate B -- Plan:
{plan_B}
Candidate B -- Code:
{code_B}

Decision rules:

Think about which candidate opens up a more promising search direction for future iterations.

Prefer the candidate that lays better groundwork for eventually achieving the best possible long-term best test score, not just the one that looks better right now.

Consider whether the approach is extensible, modular, and amenable to iterative improvement.

Prefer correctness and robustness -- a buggy candidate has no long-term potential.

Use context nodes as evidence of what has already been tried and what directions have shown promise.

Do not assume the context nodes are optimal; the new candidates may open better paths.

Output format (STRICT):

Think step by step and provide your reasoning before giving a final answer.

Give a final answer of A for Candidate A and B for Candidate B.

Provide your answer inside a \boxed{{}}, ie \boxed{{A}} or \boxed{{B}}.
\end{lstlisting}
\end{tcolorbox}
\captionof{figure}{The un-optimized Inference-only RPM prompt prior to the automated prompt optimization procedure \newline}
\label{fig:unoptimized_inference_only_prompt}

\begin{tcolorbox}[colback=md-bg, colframe=md-bg, boxrule=0pt, left=6pt, right=6pt, top=6pt, bottom=6pt, enhanced, sharp corners, breakable]
\begin{lstlisting}[basicstyle=\ttfamily\small, breaklines=true, breakatwhitespace=true, columns=fullflexible, keepspaces=true, showstringspaces=false]
You are a principal investigator allocating compute budget to one of two branches. Decide which branch is more likely to yield the best eventual test score after several iterations. Emphasize extensibility, fixability, and promise relative to what has already been tried.

Task description:
{task_desc}

Context from other solutions and their scores (not the candidates). Use this to identify promising gaps and avoid redundant directions:
{context_text}

Candidate A -- Plan:
{plan_A}
Candidate A -- Code:
{code_A}

Candidate B -- Plan:
{plan_B}
Candidate B -- Code:
{code_B}

Step-by-step evaluation:

Problem-model fit:

Does each candidate's formulation and objective align with the task? Any risks of leakage or misalignment? Note which issues are trivially fixable vs. fundamental.

Extensibility and upgrade path:

How modular is the code? How straightforward is it to add stronger models, features, or training strategies in 1 to 3 iterations?

Learning curve projection:

Based on current choices, estimate how performance might improve over the next few iterations. Identify low-hanging fruit (data cleaning, features, hyperparameters, regularization, architecture changes).

Context-informed novelty:

Relative to {context_text}, does the candidate explore a fresh, promising region or iterate intelligently on a proven one? Avoid branches that mirror underperforming context without new leverage.

Risk-adjusted potential:

Balance upside (ceiling) against effort/risk to realize it. Bugs are acceptable if the approach is sound and fixes are clear; penalize only for hard-to-remedy conceptual flaws.

Decision policy:

Choose the candidate whose direction offers higher expected long-term best test score and a credible path to get there.

Output format (STRICT):

Provide reasoning following the steps above.

End with a single final answer: A for Candidate A or B for Candidate B.

Provide your answer inside a \boxed{A} or \boxed{B}.
\end{lstlisting}
\end{tcolorbox}
\captionof{figure}{The optimized Inference-only RPM prompt after the automated prompt optimization procedure}
\label{fig:optimized_inference_only_prompt}

\subsection{Model Ensembling}
\label{app:appendix_ensembling}
We evaluate three frontier LLMs, GPT-5 \citep{singh2025openai}, Claude Opus 4.8 \citep{anthropic2026claude48} and Gemini 3.1 Pro \citep{google2026gemini31pro}, individually and ensembled, as baselines on the offline ranking evaluation detailed in Section \ref{sec:offline_eval}.

\subsubsection{Single-Model Configuration}
\label{app:frontier_model_inference_only_evals}
We leverage the same prompt produced by our prompt optimization method, and detailed in Appendix \ref{app:prompt_optimization}. Following the insights from our offline evaluations, detailed in section \ref{sec:inference_scaling}, we use the maximum reasoning effort for each model and provide the maximum number of context nodes that fit within each model's context window.

\subsubsection{Ensembling Configuration}
For both ensembling configurations, we perform 3 independent rollouts for each baseline model prior to aggregating their predictions.

\paragraph{\textbf{Majority Vote}}
This strategy employs a mechanical aggregation over the baseline models. For each rollout, the final vote of each model's response is extracted. A simple majority vote determines the final ensemble prediction, with ties broken uniformly at random.

\paragraph{\textbf{LLM Arbiter}}
The arbiter ensemble replaces mechanical vote counting with a high-level consensus call, using Claude Opus 4.8 as the final arbiter. The arbiter receives the original ranking payload (the task description, historical context, and the candidate pair) alongside the anonymized reasoning traces from the three base models. It is instructed to critically evaluate the logical quality of each expert's argument rather than blindly deferring to the majority choice. The exact template is presented in \Cref{fig:llm_arbiter_prompt}.

\begin{tcolorbox}[
    colback=md-bg,
    colframe=md-bg,
    boxrule=0pt,
    left=6pt,
    right=6pt,
    top=6pt,
    bottom=6pt,
    enhanced,
    sharp corners,
    breakable
]
\begin{lstlisting}[
    basicstyle=\ttfamily\small,
    breaklines=true,
    breakatwhitespace=true,
    columns=fullflexible,
    keepspaces=true,
    showstringspaces=false
]
{individual_ranking_prompt}

==========================
You are the FINAL ARBITER.

Above is the exact task that was given to multiple independent expert judges: a machine-learning task description, context from prior solutions, and the candidate solutions.

Each expert independently picked the candidate they believe has the higher long-term potential. Their votes and full reasoning are given below. The experts are anonymized and may disagree with one another; any individual expert may be wrong. Do NOT defer to a majority -- weigh the quality of the arguments against the task and the candidates yourself, then commit to the single best candidate.

-----------------------------------------------------------
### Expert 1
Vote: {vote_1}
Reasoning:
{reasoning_trace_1}

-----------------------------------------------------------
...
-----------------------------------------------------------
### Expert N
Vote: {vote_N}
Reasoning:
{reasoning_trace_N}
==========================
Now make the final decision among Candidate A or Candidate B.

Output format (STRICT):
- Briefly explain which expert arguments you found decisive and why.
- Give a single final answer: one letter either A or B.
- Provide your answer inside a \boxed{}, e.g. \boxed{A}.

\end{lstlisting}
\end{tcolorbox}
\captionof{figure}{LLM Ensembling Arbiter Prompt}
\label{fig:llm_arbiter_prompt}

\subsubsection{Results}
\begin{table}[t]
\centering
\begin{tabular}{l|c}
    \toprule
    \textbf{Model} & \textbf{Accuracy (\%)} \\
    \midrule
    GPT-5  & 64.66 \\
    Claude Opus 4.8 & 67.44 \\
    Gemini 3.1 Pro & 67.40 \\
    \midrule
    Ensembling (majority vote) & 68.04 \\
    Ensembling (arbiter, Claude 4.8) & \textbf{69.35} \\
    \bottomrule
\end{tabular}
\caption{Selection accuracy of the Inference-only RPM across different LLM backbones and ensembling configurations}
\label{tab:frontier-baselines}
\end{table}

Table~\ref{tab:frontier-baselines} summarizes the offline ranking results. Individually, Claude Opus 4.8 and Gemini 3.1 Pro perform nearly identically (around 67.4\%), while GPT-5 trails slightly at 64.66\%. Ensembling helps smooth out unique model errors. A standard majority vote lifts accuracy to 68.04\%, but the LLM-Arbiter achieves the top score of 69.35\%, showing that weighing the quality of arguments outperforms a simple vote count.

\subsection{Reasoning Analysis}
\label{app:reasoning_analysis}

\begin{figure}[b!]
    \centering
    \begin{minipage}{\textwidth}
    \includegraphics[width=\linewidth]{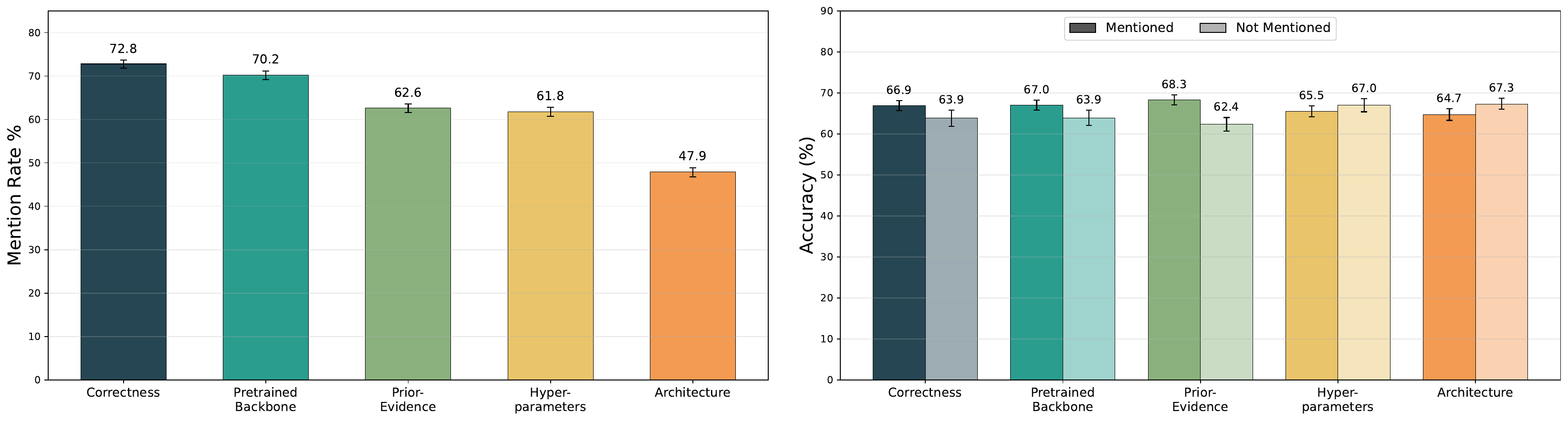}
    \end{minipage}
    \caption{\textbf{Reasoning category frequency and selection accuracy impact.} Mention frequency of reasoning categories across reasoning traces (left) and resulting selection accuracy when mentioned versus omitted (right). Citing correctness, pretrained backbones, or prior evidence improves accuracy, whereas citing architecture alone degrades performance.}
    \label{fig:inference_only_justification_categorization}
\end{figure}

\begin{figure}[t!]
    \centering
    \begin{minipage}{1.0\textwidth}
    \includegraphics[width=\linewidth]{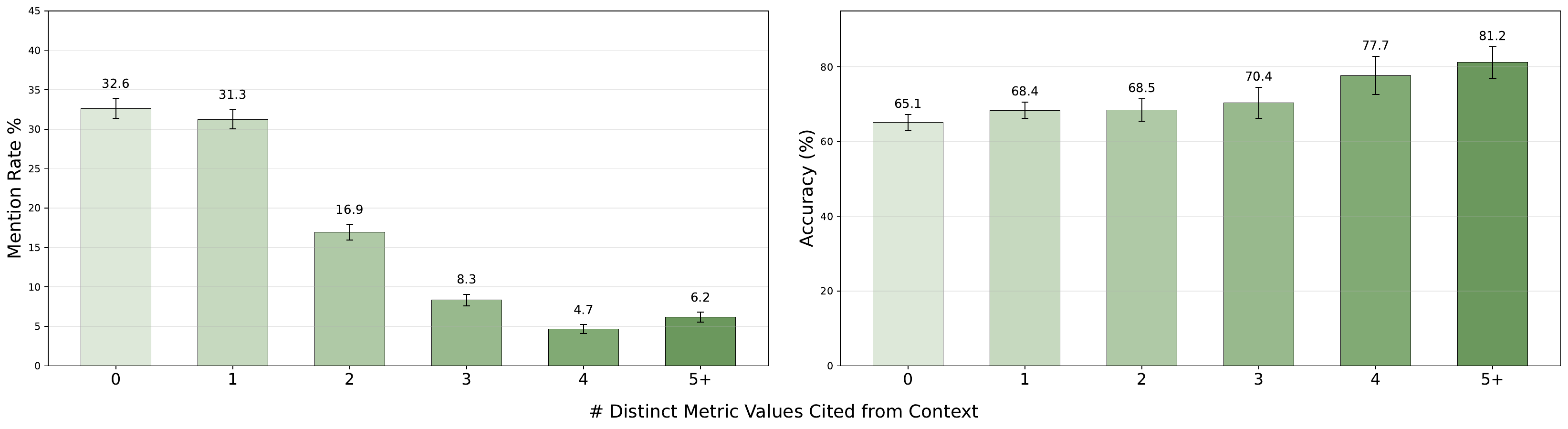}
    \end{minipage}
    \caption{\textbf{Context citation frequency and selection accuracy impact.} Mention frequency of distinct context metrics across reasoning traces (left) and resulting selection accuracy across citation counts (right). Selection accuracy increases monotonically as more context metrics are cited.}
    \label{fig:inference_only_context_citation_count}
\end{figure}

To better understand the decision-making process of inference-only RPMs, we analyze the generated reasoning traces across pairwise candidate rollouts. Specifically, we categorize the justifications used in the model's rationale and evaluate how referencing empirical context impacts selection accuracy.

\paragraph{\textbf{Evaluation Setup}}
We evaluate reasoning traces generated from 8,715 pairwise rollouts across three LLM backbones: GPT-5, Claude Opus 4.8 and Gemini 3.1 Pro, all produced as part of the evaluations in \Cref{app:frontier_model_inference_only_evals}.

\paragraph{\textbf{Justification Categorization}}
Using automated keyword matching, we categorize each reasoning trace into multi-label justification types based on five core categories:

\begin{itemize}[topsep=6pt, itemsep=3pt, parsep=0pt]
    \item \textbf{Correctness:} Identifying code bugs, syntax errors, or execution flaws.
    \item \textbf{Pretrained Backbone:} Evaluating choices of pretrained model backbones.
    \item \textbf{Prior-Evidence:} Citing measured validation/test metrics from previously explored context nodes.
    \item \textbf{Hyperparameters:} Analyzing learning rates, epoch counts, or optimizer settings.
    \item \textbf{Architecture:} Evaluating structural model modifications (e.g., fusion layers, attention blocks).
\end{itemize}

As shown in Figure~\ref{fig:inference_only_justification_categorization}, grounding selections in \emph{Correctness} ($66.9\% \text{ vs. } 63.9\%$), \emph{Pretrained Backbone} ($67.0\% \text{ vs. } 63.9\%$), or \emph{Prior-Evidence} ($68.3\% \text{ vs. } 62.4\%$) yields higher selection accuracy compared to when these justifications are omitted. Conversely, relying on abstract \emph{Architecture} justifications decreases selection accuracy from $67.3\%$ to $64.7\%$.

\paragraph{\textbf{Context Citation Analysis}}
We measure how actively the ranker utilizes historical search trajectories by counting the number of distinct context-node metric values cited in each reasoning trace. Specifically, we parse rollouts using regular expressions to extract, deduplicate, and ground explicit node identifiers and validation metrics cited from the prompt history. As shown in Figure~\ref{fig:inference_only_context_citation_count}, selection accuracy scales monotonically with the volume of cited context metrics: rollouts citing zero metric values achieve $65.1\%$ accuracy, whereas rollouts citing five or more distinct metric values reach $81.2\%$.

\newpage
\section{Agentic RPM}
\label{app:agentic_mlwm_further_details}

\subsection{Algorithm Design and Hyperparameters}
In this section we provide more implementation details for agentic RPM. As introduced in \Cref{sec:agentic_mlwm}, the core of agentic RPM is the pilot experiments. During our study, we find that the agent can be too conservative in scheduling the time budget, leaving a large portion of the time budget unused at the first call of \texttt{submit\_solution}. Even though we could prompt the  agent to run more pilot experiments after the initial submission, the follow-up experiments are often limited to hyperparameter tuning of previous ones. Over conservative time budget scheduling and repetitive follow-up experiments jointly lead to less informative pilot experiments. 

\vspace{6pt}

To deal with this problem, we use two mechanisms to elicit more informative follow-up experiments. First, we \emph{overstate} the remaining time budget in the prompt (reporting it as several times larger than it truly is), discouraging the agent from stopping prematurely. Second, after each \texttt{submit\_solution} call, a separate model reviews the findings so far and gives a \emph{feedback} that either proposes the single most informative next experiment or signals that the evidence is already sufficient, in which case we end the loop; its proposal and the remaining budget are then returned to the agent to guide the follow-up experiment. The overall workflow of agentic RPM with the overstating and feedback mechanism is presented at \Cref{alg:agentic_alg}.

\vspace{6pt}

Regarding the hyperparameters, we use the same backbone for the pilot experiment model $\pi_p$ and the feedback and final prediction model $\pi_f$ without fine-tuning.  The real time budget is $B = 300s$ even though we overstate that the time budget is $B'= 2700s$, which is $9$ times larger than the actual value. Every time a pilot experiment is finished and its empirical finding is summarized, we will terminate the loop if 1) the feedback and prediction model $\pi_f$ does not suggest the next pilot experiment; or 2) the real remaining time budget is less than a threshold $\theta = 60s$; or 3) The maximum number of pilot experiments $N=30$ is reached.  Notably, the time budget only take effect after all the tool calls in the current turn is executed. Therefore, in practice it is possible that the agentic RPM uses more time than the budget $B$ if the a tool call is extremely time-consuming.

\begin{algorithm}[H]
\begin{algorithmic}[1]
\Require Input query $q$ containing multiple candidate solutions to rank, pilot-experiment model $\pi_p$, feedback and final prediction model $\pi_f$, maximum number of pilot experiments $N$, time budget $B$, stop threshold $\theta$.
\Ensure Final decision on which candidate to choose $y_{\rm{out}}$. 
\State Initialize used time $b=0$; pilot experiment count $n=0$; pilot experiment context $x=[q]$; pilot experiment findings s = []; last submission flag $F_l= \texttt{False}$; pending experiment flag $F_p = \texttt{False}$ 
\While{True}
    \State Generate pilot-experiment model response $y_{\texttt{asst}}\sim \pi_p(\cdot\mid x)$;
    \State Parse possible tool calling $[(\texttt{tool}_i,\texttt{para}_i )]_{i=1}^T$ from response $y$, where $T$ is the number of function call
    \If {Find any $\texttt{tool}_i = \texttt{python}, i = 1,2,\ldots,T$ }
        \State Set pending experiment flag $F_p=\texttt{False}$
    \EndIf
    \If {No tools are called and $T=0$}
        \State Update  context $x = x + $ ``Please proceed to the next step using your best judgment...... ''
        \State \textbf{continue}
    \EndIf
    \State Execute the function $y^i_\texttt{tool} = \texttt{tool}_i(\texttt{para}_i),\ i=1,2,3,\ldots,T$ and update used time $b$.
    \State Update  context $x = x +  y_{\texttt{asst}} + [y^i_\texttt{tool}]_{i=1}^T $
    \If {Used time surpasses time budget $b > B$}
        \State Find $i = \mathop{\mathrm{argmax}}_i(\texttt{tool}_i =\texttt{python})$ and $j = \mathop{\mathrm{argmax}}_{j}( \texttt{tool}_j=\texttt{submit\_solution})$ 
        \If{$i>j$}
            \State Set last submission flag $F_l=\texttt{True}$
            \State Update  context $x = x + $ ``TIMEOUT. You have run out of time. Please immediately submit ...''
            \State \textbf{continue}
        \EndIf
        \State \textbf{break}
    \EndIf
    \If {$\texttt{tool}_{T} = \texttt{submit\_solution}$}
        \If {Pending experiment flag $F_p$ is $\texttt{True}$}
            \State Update $x$=$x$+ ``INVALID \texttt{submit\_solution} call...You MUST use \texttt{python} tool to run experiment''
            \State \textbf{continue}
        \EndIf
        \State $s = s+ \texttt{para}_{T}$
        \State Obtain feedback $f\sim \pi_f(\cdot\mid s)$
        \If{ The feedback $f$ does not contain next experiment}
            \State \textbf{break}
        \EndIf
        \If{ The remaining time budget is less than the stop threshold $\theta$}
            \State \textbf{break}
        \EndIf
        \State Set the pending experiment flag $F_p$ to be $\texttt{True}$.
        \State Update context $x = x + f$
        \If {$F_l$ is $\texttt{True}$ }
            \State \textbf{break}
        \EndIf
        \If {The maximum number of pilot experiments is reached $\texttt{len}(s) \geq N$ }
            \State \textbf{break}
        \EndIf
    \EndIf
    \If {The last submission flag $F_l$ is \texttt{True}}
        \State \textbf{break}
    \EndIf
\EndWhile
\State Make the final decision $y_{\rm{out}}\sim \pi_f(\cdot\mid s)$
\State {\bfseries Return: } Final decision $y_{\rm{out}}$
\end{algorithmic}
\caption{The workflow of the proposed agentic RPM.}
\label{alg:agentic_alg}
\end{algorithm}

\newpage

\subsection{Design Ablations}
Both end-to-end (\Cref{sec:e2e_eval}) and offline evaluations (\Cref{sec:agentic_rpm_offline_eval_results}) demonstrate the efficacy of the Agentic RPM to accelerate and enhance AIRA. To have a better understanding of how different components contribute, in this section we conduct an ablation study on the proposed Agentic RPM. Specifically, we experiment with the following variants: 
\begin{itemize}[topsep=6pt, itemsep=3pt, parsep=0pt]
    \item \textbf{w/o overstating}: The agent is informed with the real remaining time budget without any exaggeration;
    \item \textbf{w/o feedback}: The feedback mechanism is removed and the it receives no planning about the next valuable experiment to run. 
    \item \textbf{w/o GPU access}: The agent no longer has access to the H200 GPU and thus can only run pilot experiments with a CPU. 
\end{itemize}

\begin{figure}[b!]
    \centering
    \begin{minipage}{0.24\textwidth}
        \includegraphics[width=\linewidth]{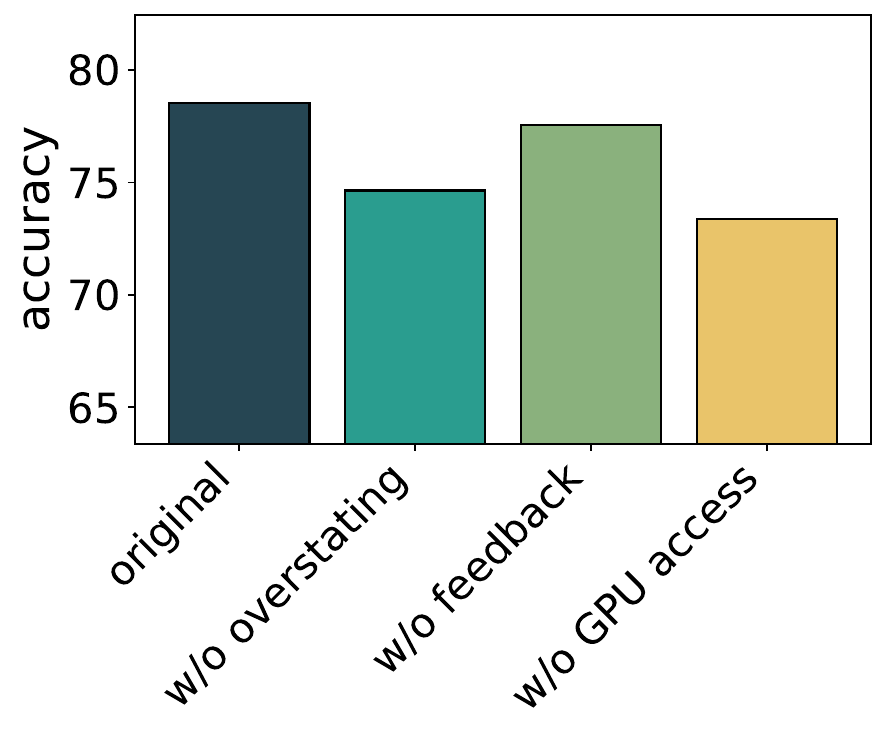}
    \end{minipage}
    \begin{minipage}{0.24\textwidth}
        \includegraphics[width=1.0\linewidth]{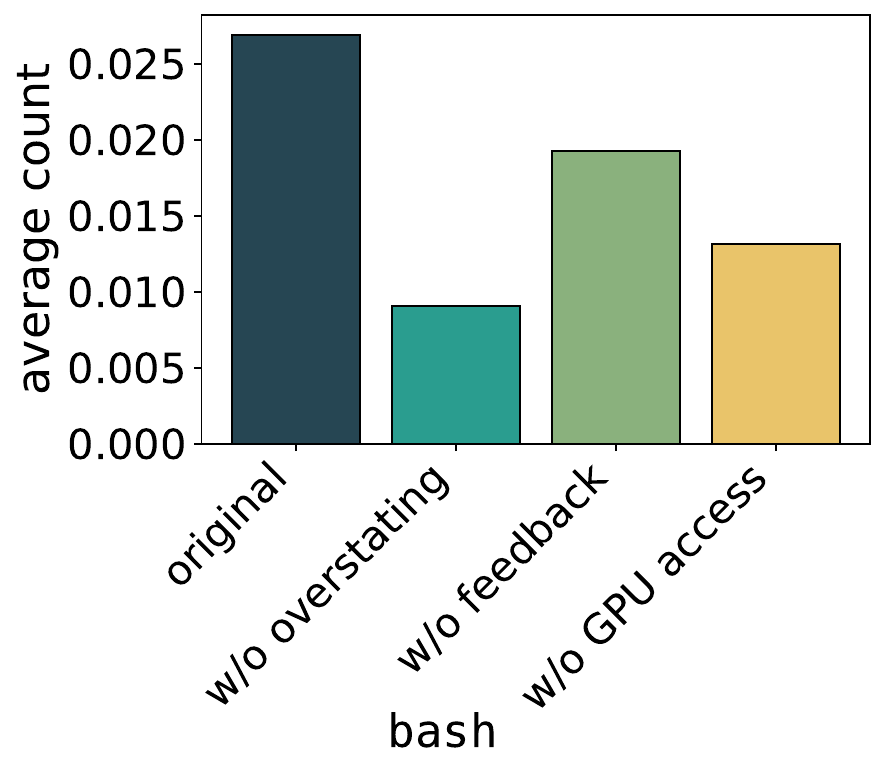}
    \end{minipage}
    \begin{minipage}{0.24\textwidth}
        \includegraphics[width=1.0\linewidth]{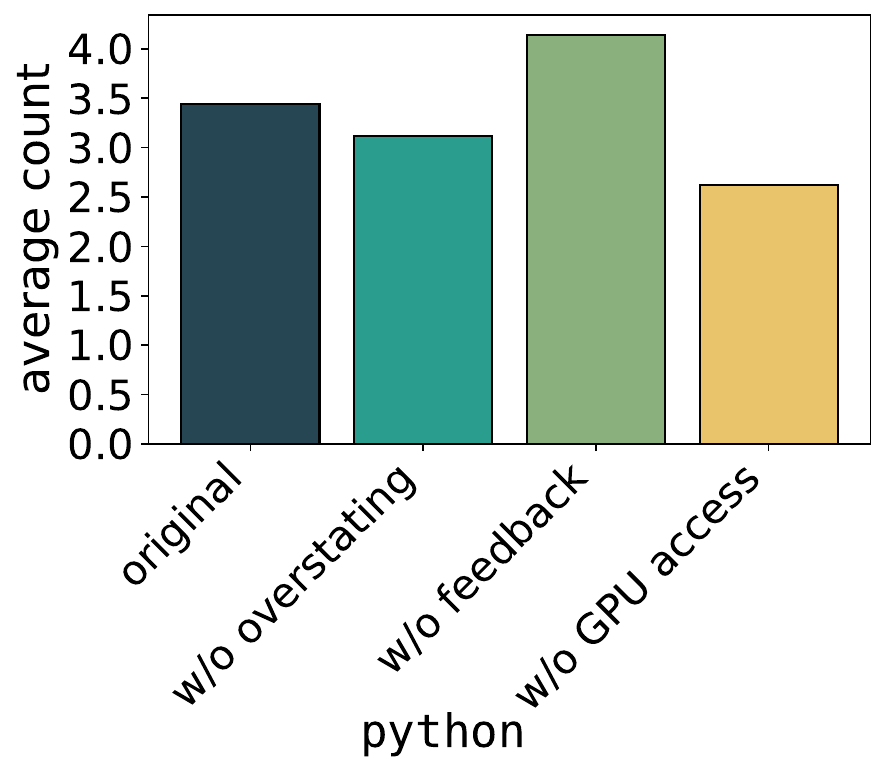}
    \end{minipage}
    \begin{minipage}{0.24\textwidth}
        \includegraphics[width=1.0\linewidth]{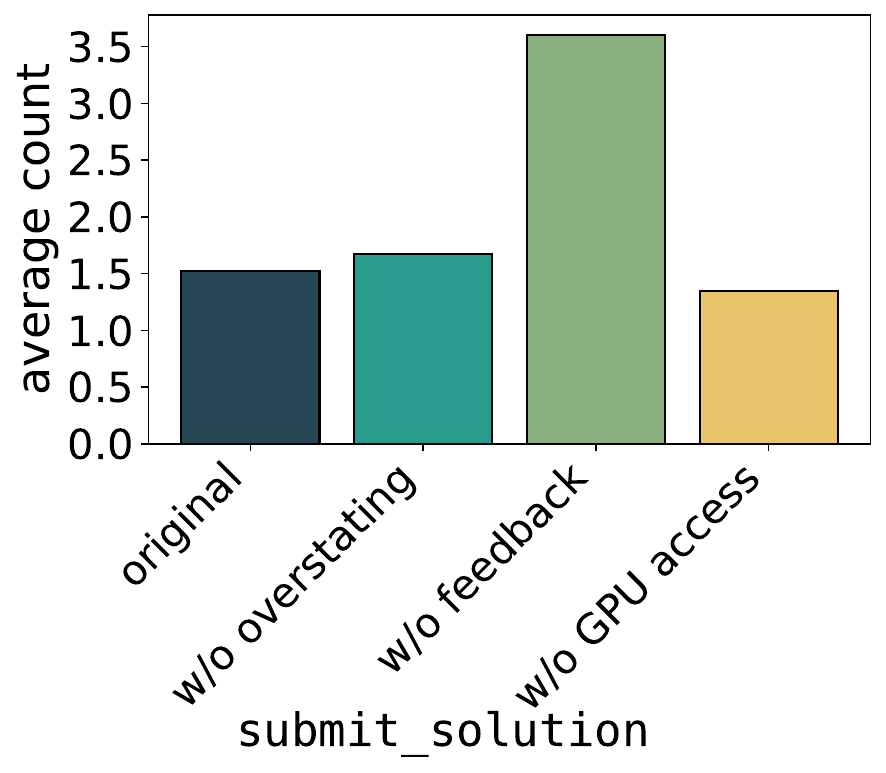}
    \end{minipage}
    \caption{The performance of three ablation variants at offline evaluation using GPT-5 (leftmost) and their average tool calling count of \texttt{bash} (middle left), \texttt{python} (middle right), \texttt{submit\_solution} (rightmost). \texttt{bash} is seldom used by any variant. The \emph{w/o feedback} variant calls \texttt{python} and \texttt{submit\_solution} more frequently than other variants.}
    \label{fig:agentic_ablation_and_tool_use}
\end{figure}

\begin{figure}[b!]
    \centering
    \begin{minipage}{0.32\textwidth}
        \includegraphics[width=\linewidth]{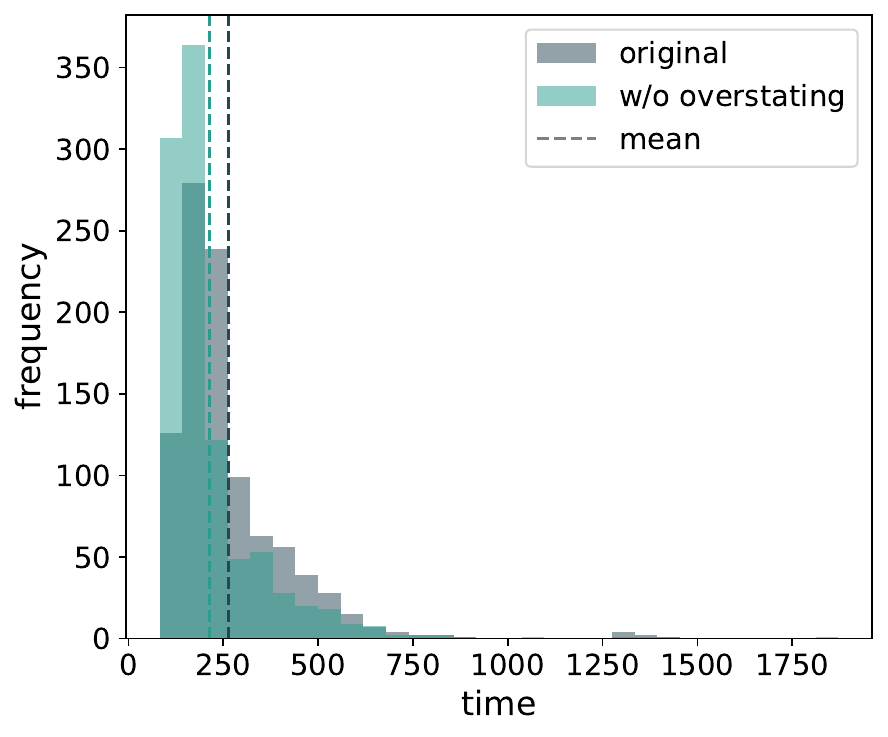}
        % \subcaption{The average execution time distribution of original agentic RPM and w/o overstating variant.}
    \end{minipage}
    \hfill
    \begin{minipage}{0.32\textwidth}
        \includegraphics[width=1.0\linewidth]{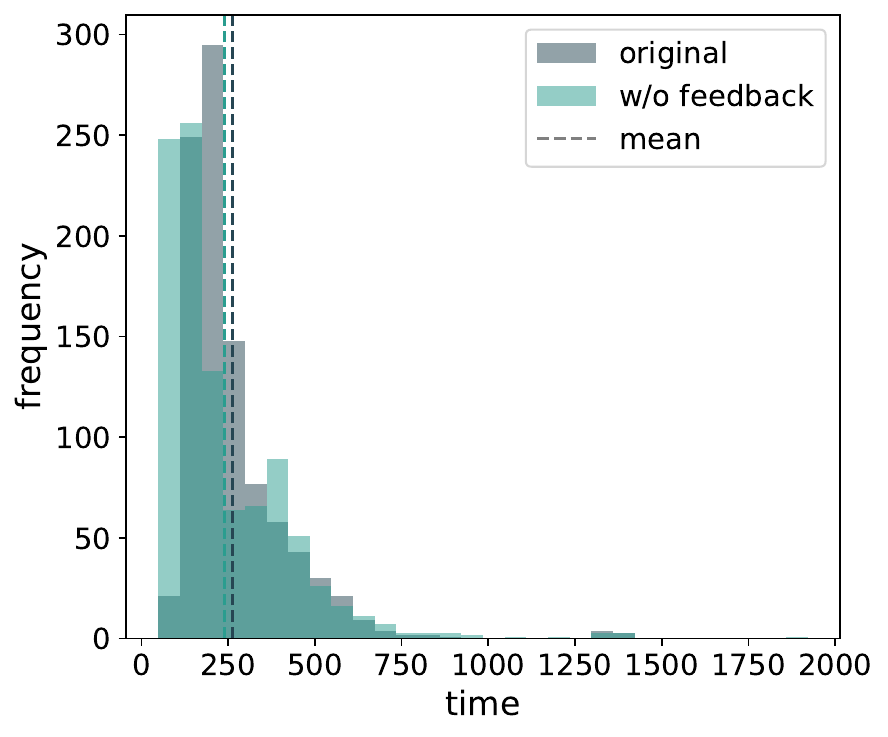}
        % \subcaption{The average execution time distribution of original agentic RPM and w/o feedback variant.}
    \end{minipage}
    \hfill
    \begin{minipage}{0.32\textwidth}
        \includegraphics[width=1.0\linewidth]{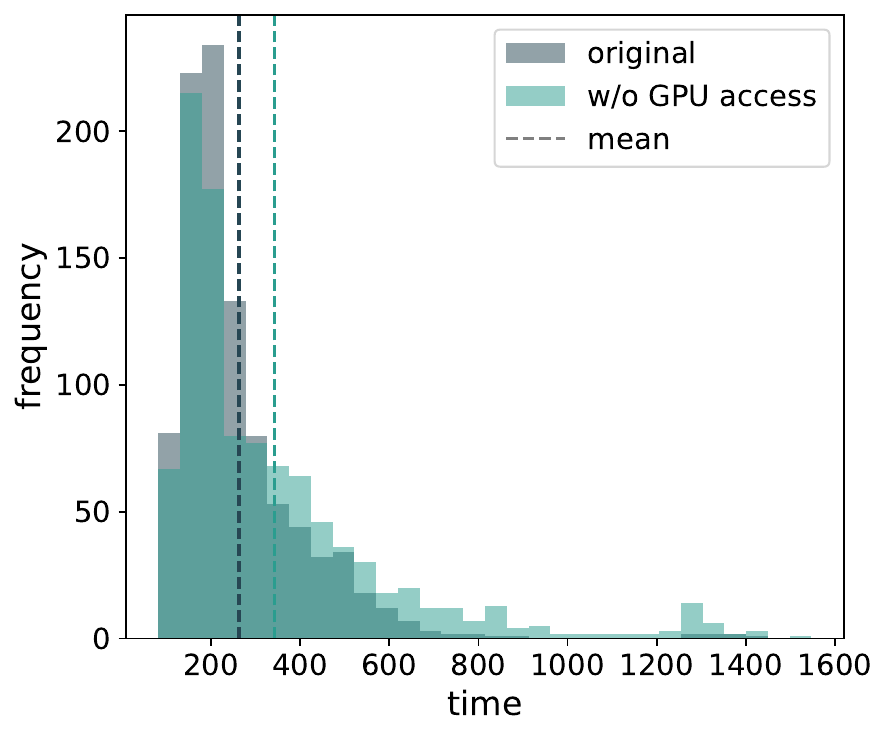}
        % \subcaption{The average execution time distribution of original agentic RPM and w/o GPU access variant.}
    \end{minipage}
    \caption{The average execution time distribution of pilot experiments across offline evaluations. Compared with \emph{w/o overstating} variant (left) and \emph{w/o feedback} variant (middle), the original Agentic RPM produces more time-consuming pilot experiments. Without GPU access (right), the average execution time is further increased. }
    \label{fig:avg_elapsed_per_dp}
\end{figure}

We follow the same offline evaluation setup presented in \Cref{sec:offline_eval}, and provide the Agentic RPM a 5-minute compute time budget. As presented in \Cref{fig:agentic_ablation_and_tool_use}, the inclusion of the feedback mechanims and overstating the provides time budget provides an improvement in selection accuracy. We also observe a significant drop in performance when removing GPU access for the pilot experiments. To further understand the behavior difference between the original  agent and its variants, we analyze the average tool call count and the distribution of the pilot-experiments' running time. The results are presented in \Cref{fig:agentic_ablation_and_tool_use} and \Cref{fig:avg_elapsed_per_dp} respectively. From the figures we can observe that: (1) removing time budget overstating leads to less \texttt{python} call and less time-consuming experiments; (2) removing the GPU access increases the time cost of each experiment since the pilot experiments can only be run on CPU in this case; (3) removing feedback mechanism leads to more and briefer experiment, suggesting that the pilot experiments it runs can be trivial and less informative.  

\subsection{Prompt Templates}

In this section, we detail the complete set of prompts guiding the agentic RPM workflow. Specifically, we present the prompt templates to initialize the workflow (Figure~\ref{fig:arq_prompt}), to provide  feedback (Figure~\ref{fig:arq_feedback_prompt}), and to give a final selection (Figure~\ref{fig:arq_final_prompt}).

\begin{tcolorbox}[
    width=1\linewidth,
    colback=gray!7,
    colframe=black!0,
    title=Prompt for Stepping Stone Generation,
    fonttitle=\bfseries,
    boxrule=0.5pt,
    breakable,
]
\small

\textbf{INTRODUCTION}:

You are a Kaggle Grandmaster and Lead Data Scientist acting as a \textbf{Strategic Evaluator}. Your task is to analyze two candidate machine learning solutions and \textbf{predict which one will achieve a superior test score without actually evaluating on the test set}.\\
You are given two candidate solutions for a machine learning competition on Kaggle. We do not have access to the test set. Instead, we estimate test performance by running 5-fold cross-validation (CV) on the public training set. However, running the full 5-fold CV on the entire training set is prohibitively time-consuming and beyond our computational budget. Therefore, you must predict which solution would achieve a better test score if it were fully executed, using fast, high-leverage experiments (Pilot Experiments) on the public training set.\\
Remember: your goal is NOT to build the final solution, but to generate code that helps \textbf{predict the relative ranking of the two solutions}.\\

\textbf{TASK DESCRIPTION}
\begin{verbatim} 
```
{task_desc}
```
\end{verbatim}

You can access the data using standard Python libraries (pandas, datasets, etc.).\\
The data is already mounted and ready to use — no need to download or prepare it.\\
\textbf{Important:} Pay attention to whether higher or lower values indicate better performance (e.g., accuracy is higher-is-better, and RMSE is lower-is-better).\\
\textbf{Common Pitfalls:} Do NOT use \texttt{from transformers import AdamW} (it has been removed). Use \texttt{from torch.optim import AdamW} instead.\\
\textbf{Your Resources (as Strategy Evaluator):} \{\texttt{device}\}\\
\textbf{Time Limit:} \{\texttt{time\_limit}\}\\

\textbf{CANDIDATE SOLUTIONS:} \\
\textbf{Option A:}\begin{verbatim}
```python
{candidate_a}
```
\end{verbatim}

\textbf{Option B:}\begin{verbatim}
```python
{candidate_b}
```
\end{verbatim}

......\\

\textbf{INSTRUCTIONS FOR PILOT EXPERIMENTS:}\\
You must write a \textbf{fast, efficient Python script} (with a target runtime under \texttt{\{time\_limit\}}). When executed, this script should produce evidence that helps determine which candidate solution would achieve a better test score.\\

You have full freedom in choosing your analysis strategy. For example, you may:\\
- Run a simplified or partial cross-validation (e.g., fewer folds, subsampled data).\\
- Compare key architectural or hyperparameter differences between the two solutions and run targeted ablation experiments.\\
- Evaluate both solutions on a single train/validation split.\\
- Perform any other analysis you believe is informative.\\

\textbf{REQUIREMENT}:\\
Step 1: Use the above information and the provided tools to first understand the task and the candidate solutions, and then write a Python script to generate evidence.\\
Step 2: Use the provided tools to fix any bugs in your script and run it, adding any additional experiments needed to strengthen the evidence.\\
Step 3: You \textbf{MUST} summarize your experiment design rationale, experimental results and findings in plain text and submit them with the tool \texttt{submit\_solution()}.

\end{tcolorbox}
\captionof{figure}{Prompt used for initializing the agentic RPM. The \texttt{task\_desc}, \texttt{device}, \texttt{time\_limit}, \texttt{candidate\_a} and \texttt{candidate\_b} are placeholders.}
\label{fig:arq_prompt}

\begin{tcolorbox}[
    width=1\linewidth,
    colback=gray!7,
    colframe=white!0,
    title=Prompt for Stepping Stone Generation,
    % fonttitle=\bfseries,
    % boxrule=0.5pt,
    breakable,
]
\small

\textbf{INTRODUCTION}:

You are a Kaggle Grandmaster and Lead Data Scientist acting as an \textbf{Experiment Planner}. Your task is to decide whether more pilot experimentation is needed, and if so, propose the \textbf{single most valuable next experiment} for distinguishing which of the candidate machine learning solutions is most likely to achieve the best test score.

You are given a set of candidate solutions for a machine learning competition on Kaggle. You only have access to the public training set but do not have access to the test set. A natural proxy for test performance would be 5-fold cross-validation (CV) score on the public training set. However, running the full 5-fold CV on the entire training set is time-consuming. Therefore, your goal is to design diverse pilot experiments that can approximate relative test performance \textbf{with significantly less computation} than standard 5-fold CV.\\

\textbf{Important:} Your job is \textbf{only} to decide whether to stop or to design the \textbf{next best experiment}. A separate coding agent will execute the experiment and collect results, and a separate judging agent will make the final prediction based on the experimental evidence. You should NOT write any code yourself, and you should NOT make the final prediction yourself.\\

\textbf{TASK DESCRIPTION}\begin{verbatim}
```
{task_desc}
```
\end{verbatim}

\textbf{ENVIRONMENT \& CONSTRAINTS (for the coding agent to follow)}:\\
\textbf{Data Access:} You can access the data using standard Python libraries (pandas, datasets, etc.). The data is already mounted and ready to use — no need to download or prepare it.\\
\textbf{Metric:} Pay attention to whether higher or lower values indicate better performance (e.g., accuracy is higher-is-better, RMSE is lower-is-better).\\
\textbf{Common Pitfalls:} Do NOT use \texttt{from transformers import AdamW} (it has been removed). 
Use \texttt{from torch.optim import AdamW} instead.\\
\textbf{Computation Resources:} \{device\}\\
\textbf{Time Limit:} \{time\_limit\}.  If you propose a next experiment, it must fit within the remaining time budget with room left for execution reporting and a possible final decision.

\textbf{CANDIDATE SOLUTIONS}:\\
\textbf{Option A:}\begin{verbatim}
```python
{candidate_a}
```
\end{verbatim}

\textbf{Option B:}\begin{verbatim}
```python
{candidate_b}
```
\end{verbatim}

......\\

\{prev\_findings\}\\

\textbf{INSTRUCTIONS}:

You must first decide whether the existing evidence is already sufficient. If yes, recommend stopping. If not, propose exactly \textbf{one} next experiment that is expected to provide the highest decision value.\\
The next experiment should be detailed and specific enough that a coding agent can implement and execute it by strictly following your instructions.\\
Prefer experiments that reduce the most important unresolved uncertainty about the relative ranking of the candidate options (Option A, Option B, Option C, ...).\\
Do NOT propose a trivial repetition of an earlier experiment unless you clearly justify why a stability check is necessary.\\
Do NOT propose cosmetic variations of previous experiments (for example, the same proxy, same split logic, and same reasoning with only a tiny tweak) unless that variation is specifically needed to resolve an important uncertainty.\\
Prefer experiments whose outcome could realistically change the current provisional ranking.\\
If the existing evidence is already strong enough and another experiment is unlikely to add meaningful value, set `stop: true`.\\

You have full freedom in designing your strategy. For example, you may:\\
- Run a simplified or partial cross-validation (e.g., fewer folds, subsampled data).\\
- Compare key architectural or hyperparameter differences among the candidate solutions and run targeted ablation experiments.\\
- Evaluate the candidate solutions on a single train/validation split.\\
- Perform any other analysis you believe is informative.\\

If you propose a next experiment, it must include:\\

- \textbf{title} (str): A short descriptive name for the experiment.\\
- \textbf{goal} (str): What this experiment is trying to find out.\\
- \textbf{steps} (str): A detailed, step-by-step description of what the coding agent should implement and run.\\
- \textbf{expected\_runtime} (float): A rough estimate of how long this experiment will take in minutes given the computation resources.\\

\textbf{OUTPUT REQUIREMENT}:
Output a JSON object with the following schema:
\begin{verbatim}
```json
{{
  "stop": <true_or_false>,
  "reason": "<short explanation>",
  "next_experiment": {{
    "title": "<string>",
    "goal": "<string>",
    "steps": "<string>",
    "expected_runtime": <float>
  }}
}}
```
\end{verbatim}

If "stop": true,  then set:
 \begin{verbatim}
```json
"next_experiment": null
```
\end{verbatim}
Important:\\
- Output strictly valid JSON only.\\
- Do not wrap the JSON in Markdown fences.\\
- All string values must be valid JSON strings with double quotes.\\
- "steps" should be a single string. You may use `\\n` inside the string for line breaks.\\

Example Output:
\begin{verbatim}

{{
  "stop": false,
  "reason": "Existing micro-split results are mixed and high-variance. A decisive equal-compute comparison 
   is still needed.",
  "next_experiment": {{
    "title": "Matched-step class-balanced holdout",
    "goal": "Run a matched-step comparison between the candidate options under the same compute budget.",
    "steps": "1. Construct a class-balanced holdout split.
              2. Match training steps and optimizer settings.
              3. Run each candidate option under identical budget.
              4. Compare validation accuracy.",
    "expected_runtime": 6.0
  }}
}}
"""
\end{verbatim}
\end{tcolorbox}

\captionof{figure}{Prompt used to provide feedback in agentic RPM workflow. The \texttt{task\_desc}, \texttt{device}, \texttt{time\_limit}, \texttt{candidate\_a}, \texttt{candidate\_b} and \texttt{prev\_findings} are placeholders.}
\label{fig:arq_feedback_prompt}

\begin{tcolorbox}[
    width=1\linewidth,
    colback=gray!7,
    colframe=white!0,
    title=Prompt for Stepping Stone Generation,
    % fonttitle=\bfseries,
    % boxrule=0.5pt,
    breakable,
]
\small

\textbf{INTRODUCTION}:

You are a Kaggle Grandmaster and Lead Data Scientist acting as a \textbf{Final Prediction Analyst}.
Your task is to analyze multiple machine learning solutions and their pilot experiment results to predict which solution will have superior test cross-validation performance.

You have already conducted fast pilot experiments to probe the potential of these multiple solutions. Now you must analyze the empirical evidence from these experiments and make a final prediction.

\textbf{TASK DESCRIPTION}
\begin{verbatim}
```
{task_desc}
```
\end{verbatim}

\textbf{CANDIDATE SOLUTIONS}:

\textbf{Option A:}\begin{verbatim}
```python
{candidate_a}
```
\end{verbatim}

\textbf{Option B:}\begin{verbatim}
```python
{candidate_b}
```
\end{verbatim}

\texttt{\{execution\_output\}}

\textbf{INSTRUCTIONS}:

Based on the pilot experiment results above, you must predict which candidate solution will achieve the best test cross-validation score when fully implemented.\\

Consider the following in your analysis:\\
- \textbf{Empirical Evidence:} What do the execution outputs tell you about the performance of each solution?\\
- \textbf{Reliability:} Are the results consistent and reliable, or are there signs of instability?\\
- \textbf{Potential:} Which solution shows more promise for achieving higher validation metrics?\\
- \textbf{Implementation Quality:} Does the output suggest successful execution or potential issues?\\

\textbf{OUTPUT REQUIREMENT}:

Think step by step and provide your reasoning before giving a final answer.
Each candidate solution above is labeled with a letter (Option A, Option B, Option C, ...). Your final answer must be the single letter of the best option, enclosed in \texttt{\textbackslash\textbackslash boxed\{\{\}\}} — for example \texttt{\textbackslash\textbackslash boxed\{\{A\}\}}, \texttt{\textbackslash\textbackslash boxed\{\{B\}\}}, or \texttt{\textbackslash\textbackslash boxed\{\{C\}\}}. Output exactly one letter, and only choose from the options listed above.\\

Example response format:

\textbf{Analysis}: [Your detailed reasoning here] \\

\textbf{Final Prediction}: \texttt{\textbackslash\textbackslash boxed\{\{A\}\}}

\end{tcolorbox}
\captionof{figure}{Prompt used for the final prediction in agentic RPM workflow. The \texttt{task\_desc},  \texttt{candidate\_a}, \texttt{candidate\_b} and \texttt{execution\_output} are placeholders.}
\label{fig:arq_final_prompt}

% \subsection{Ablation Study}

\newpage
\section{End-to-End Evaluations}
\subsection{Impact of Selection Quality}
\label{app:selection_quality_e2e_impact}

To determine whether local decision quality directly drives downstream agent performance, we retrospectively analyze the relationship between step-wise candidate selection quality and final end-to-end search scores.

\begin{figure}[h]
    \centering
    \begin{minipage}{0.65\textwidth}
    \includegraphics[width=\linewidth]{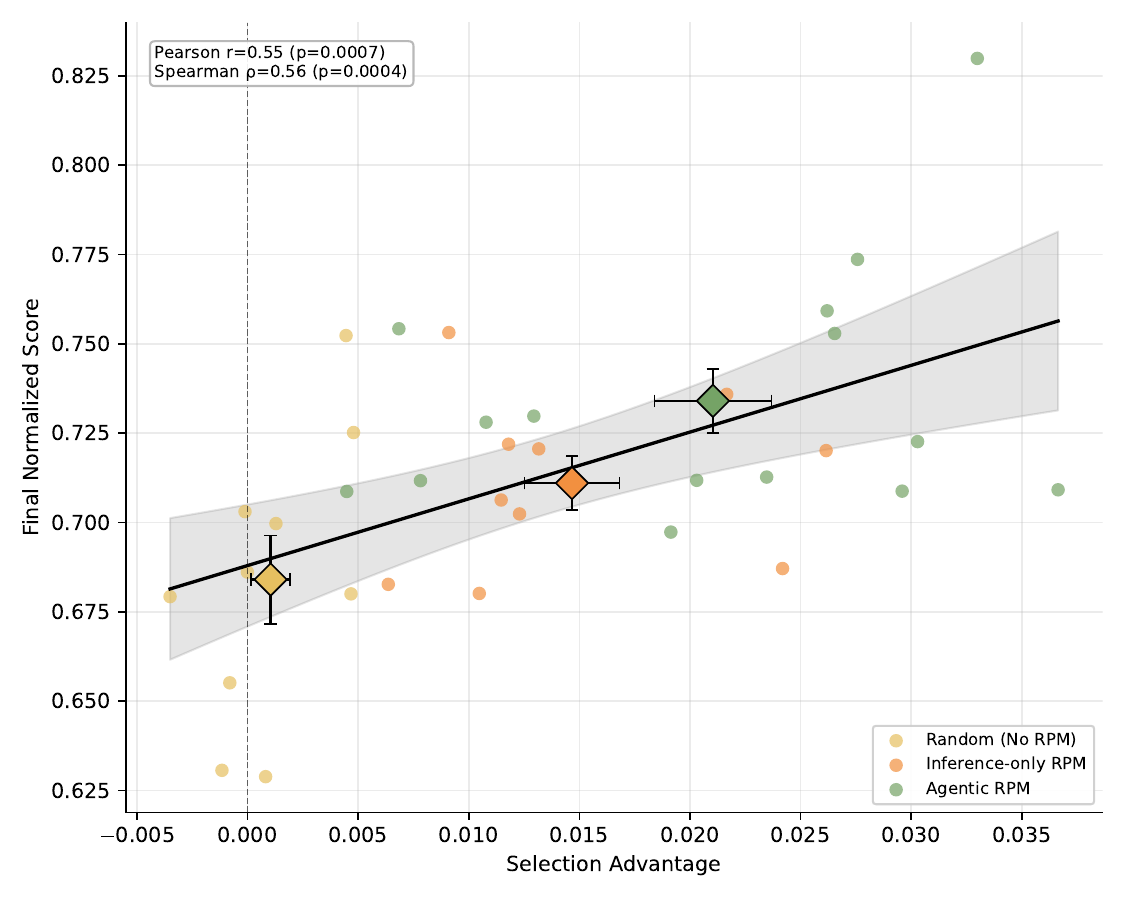}
    \end{minipage}
    \caption{\textbf{Selection Advantage vs. Final Normalized Score.} Final normalized score plotted against step-wise selection advantage across individual seed runs. Large diamonds denote mean values ($\pm 95\%$ CI). Selection advantage strongly correlates with end-to-end performance (Pearson $r=0.55, p=0.0007$; Spearman $\rho=0.56, p=0.0004$), with Agentic RPM achieving both higher selection advantage and final score than Inference-only RPM and Random selection.}
    \label{fig:selection_quality_e2e_impact}
\end{figure}

\paragraph{\textbf{Evaluation Setup}}
Each data point in Figure~\ref{fig:selection_quality_e2e_impact} represents an individual seed run, where the final score is the normalized metric averaged across all AIRS-Bench tasks. To quantify decision quality, we retrospectively evaluate all $N$ candidate solutions generated at each step against ground-truth benchmarks. We define \emph{Selection Advantage} as $S_{\text{chosen}} - \frac{1}{N} \sum_{i=1}^{N} S_i$, measuring the difference between the chosen candidate's ground-truth score $S_{\text{chosen}}$ and the mean score of all $N$ candidates in the candidate pool. For each seed run, we report the selection advantage averaged across all selection steps and across all tasks.

\paragraph{\textbf{Findings}}
We find a statistically significant positive correlation between the selection advantage and the final end-to-end performance (Pearson $r=0.55, p=0.0007$; Spearman $\rho=0.56, p=0.0004$). As expected, random selection yields an average selection advantage near zero. Both preference models consistently improve selection quality over random selection, with the Agentic RPM achieving the highest overall selection advantage and final score. These results confirm that higher per-step candidate selection quality translates directly to better end-to-end AIRA search performance.

% =====================================================================
%  Pick Your Poison: Does an MLWM's code judgment add signal
%  beyond the validation-score oracle?
%
%  Contribution to "Searching for Signal: Machine Learning World
%  Models for AI Research Agents" (MLWM paper).
%  Author: Bhavul Gauri.  Methodology: Pick-Your-Poison offline eval.
%  (Prompt lstlisting style `prompt` is defined in sections/appendix.tex.)
% =====================================================================
\newpage
\section{RPM-Augmented Final Node Selection}
\label{app:pickyourpoison}

%% Sub-Sectionless Intro - Overview about what differentiates

As presented in Sections \ref{sec:scaffolds} and \ref{sec:aira_mlwm_integration}, an RPM could intervene at three points within an evolutionary search based AIRA's search: parent selection (PS), child creation (CC) and final-node selection (FNS).

\vspace{6pt}

A key distinction separates CC from PS and FNS: CC operates on unexecuted candidates at selection time, while PS and FNS candidates have already been executed and carry validation scores. Thus, a strong selection heuristic (such as greedy selection) can be trivially constructed for PS and FNS. As such, we focus in this manuscript on augmenting CC within an AIRA, but present exploratory initial results on augmenting FNS.

%% Evaluation Method Validation and Test + Prompt
\subsection{Evaluation Setup}
\label{sec:pyp_eval_setup}
To evaluate the capability of RPMs to perform FNS, we utilize the search trees produced by the Inference-only RPM End-to-End evaluations across the public text-and-tabular \airsbench{} tasks (\Cref{sec:e2e_eval_setup,sec:e2e_eval}). Each datapoint in the evaluation represents a complete \airadojo{} search tree on a specific task, and the RPM is tasked with selecting the candidate solution with the highest test score from this tree. We use Qwen3.6-27B \citep{qwen3.6-27b} as the LLM backbone for FNS to maintain model consistency between the original \airadojo{} operators and selection. We also evaluate GPT-5 as an alternative backbone to validate how stronger selection quality impacts FNS.

\vspace{6pt}

To contextualize overall performance, we compare against two global reference baselines: a global \emph{Test Oracle} (the maximum test score in the entire search tree) and a global \emph{Validation Oracle} (the test score of the candidate with the overall highest validation score; the default FNS mechanism of \airadojo{}). Performance is measured using the average normalized test score of the selected candidate.

\subsection{Method}
\label{sec:pyp_method}
Given an \airadojo{} search tree, we filter out buggy nodes and retain all valid candidate solutions, with validation and test metrics normalized following \Cref{sec:aira_benchmarks}. Because an \airadojo{} search tree can contain hundreds of nodes, we restrict the FNS candidate pool to the $N$ nodes with the highest validation scores, sweeping across pool sizes $N \in \{2, 4, 6, 8, 10\}$. To contextualize performance within each candidate pool, we evaluate against two local baselines: a local \emph{Test Oracle ($N$ shown)} representing the best test score available within the $N$-candidate pool, and a \emph{Random ($N$ shown)} baseline representing the expected score of a uniform random pick from that pool.

\vspace{6pt}

Selection from the candidate pool is conducted via a round-robin tournament, where across 30 repeated matches, candidate subsets of $\min(N,5)$ are presented to the Inference-only RPM to select between. The candidate accumulating the highest point total across all matches is selected as the tournament winner, with ties broken uniformly at random. The complete prompt template used for the Inference-only RPM during tournament matches is detailed in \Cref{fig:pyp_prompt}.

\begin{tcolorbox}[colback=md-bg, colframe=md-bg, boxrule=0pt, left=6pt, right=6pt, top=6pt, bottom=6pt, enhanced, sharp corners, breakable]
\begin{lstlisting}[basicstyle=\ttfamily\small, breaklines=true, breakatwhitespace=true, columns=fullflexible, keepspaces=true, showstringspaces=false]
You are a strict judge selecting the BEST among 5 candidate solutions to the SAME machine learning task.

Your goal is to choose the candidate most likely to achieve a better test score.

Task description:
```markdown
{{task_description}}
```

Context from various solutions to the same machine learning task. These are NOT the candidates you are judging.
{{context_nodes_description}}

Candidate A - normalized_validation_metric: {{candidate_a_score}}
Candidate A - Code:
```python
{{candidate_a_code}}
```

Candidate B - normalized_validation_metric: {{candidate_b_score}}
Candidate B - Code:
```python
{{candidate_b_code}}
```

Candidate C - normalized_validation_metric: {{candidate_c_score}}
Candidate C - Code:
```python
{{candidate_c_code}}
```

Candidate D - normalized_validation_metric: {{candidate_d_score}}
Candidate D - Code:
```python
{{candidate_d_code}}
```

Candidate E - normalized_validation_metric: {{candidate_e_score}}
Candidate E - Code:
```python
{{candidate_e_code}}
```

Decision rules:
- Prefer the candidate most likely to produce a better test score for the given task.
- Prefer correctness, robustness, and task-fit over style or verbosity.
- Use context nodes only as supporting evidence (e.g., what has already been tried, what validation score looked like).
- Do not assume the context nodes are optimal; the new candidates may be better.

Output format (STRICT):
- Think step by step and provide your reasoning before giving a final answer.
- Give a final answer of A, B, C, D or E.
- Provide your answer inside a \boxed{}, ie \boxed{A}, \boxed{B}, \boxed{C}, \boxed{D}, \boxed{E}.
\end{lstlisting}
\end{tcolorbox}
\captionof{figure}{The Inference-only RPM's prompt for final-node selection. For pool sizes $N$ where $N<5$, the match size is capped at $N$ and the boxed-letter menu shrinks accordingly.}
\label{fig:pyp_prompt}

%% Results (validation and test)

\subsection{Results}
\label{sec:pyp_results}

\begin{figure}[t!]
    \centering
    \includegraphics[width=0.6\linewidth]{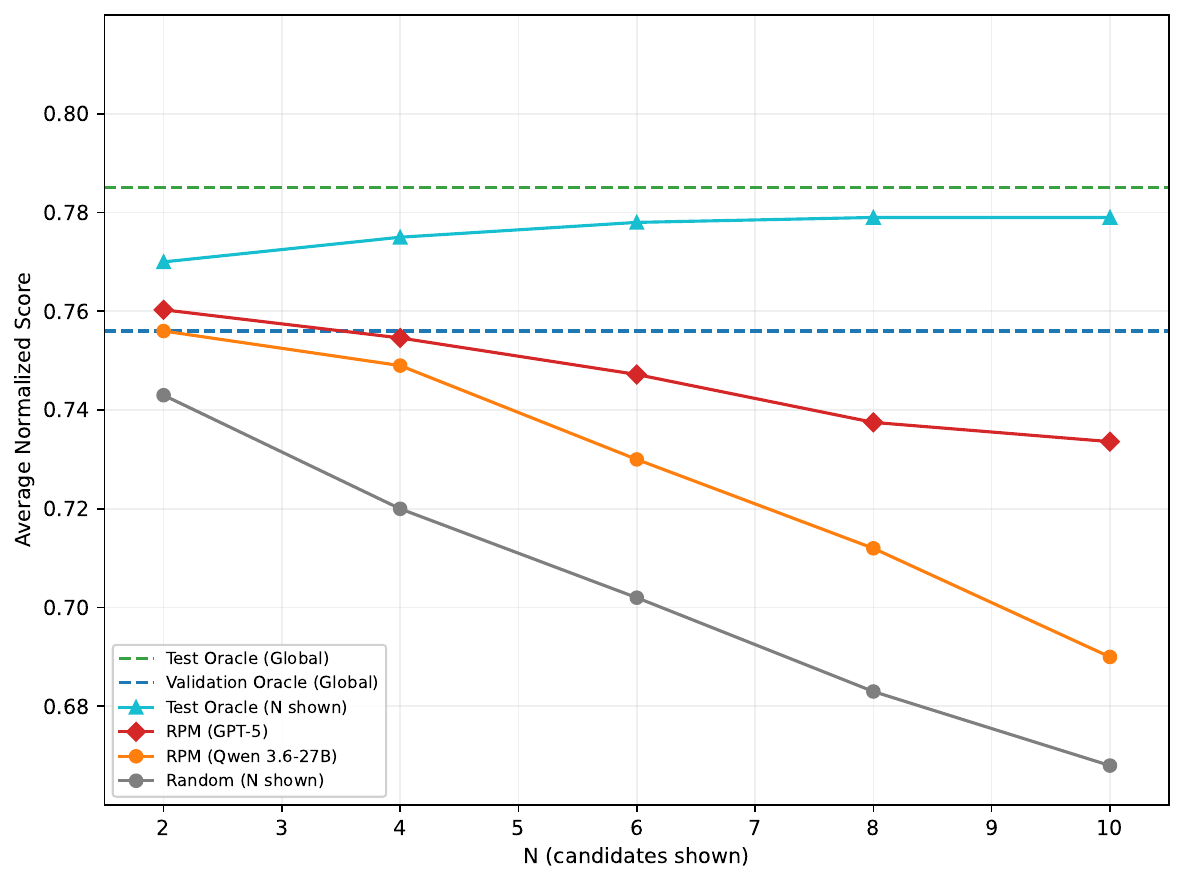}
    \caption{\textbf{Final Node Selection performance across candidate pool sizes.} Average normalized test score as a function of the top-$N$ validation candidate pool size. Inference-only RPM judges (GPT-5 and Qwen3.6-27B) beat random selection but degrade as $N$ increases, becoming progressively worse than the Validation Oracle (\airadojo 's default selection strategy). The Inference-only RPM with the stronger GPT-5 LLM backbone outperforms that with the Qwen backbone.}
    \label{fig:pyp_fns}
\end{figure}

As shown in \Cref{fig:pyp_fns}, both RPM variants consistently outperform the local random baseline across all pool sizes $N$. Leveraging GPT-5 yields consistently higher normalized test scores than Qwen3.6-27B, confirming that stronger base reasoning directly improves selection quality. However, the performance of both RPMs degrades as $N$ increases, sliding further below the global \emph{Validation Oracle} (\airadojo 's default selection mechanism) as the judge attempts to separate between an increasingly weaker pool of candidates.

\vspace{6pt}

These results illustrate the fundamental challenge of augmenting FNS: once candidates are executed, validation-based selection enables strong heuristics that are difficult to improve upon purely through code inspection. This is further limited given the Hidden Consistent Evaluation protocol's strong test-validation generalization, the gap between the \emph{Validation Oracle} and the ultimate \emph{Test Oracle} is extremely narrow. While static inference-only judgment struggles to outperform this greedy baseline on executed nodes, future work could deploy more active selection paradigms, such as the Agentic RPM, to run targeted experiments and bridge the remaining gap to the Test Oracle.

\end{document}